\documentclass[journal]{IEEEtran}
\usepackage{graphicx}
\usepackage{amssymb}
\usepackage{amsmath}
\usepackage{multirow}
\usepackage{rotating}
\usepackage{colortbl}
\usepackage[table]{xcolor}
\usepackage{caption}
\usepackage{hyperref}

\begin{document}
%
\title{Joint-Space Control of a Structurally\\Elastic Humanoid Robot}
%
%
%

\author{Connor~W.~Herron*,~Christian~Runyon,~Isaac~Pressgrove,~Benjamin~C.~Beiter,~Bhaben~Kalita,\\~and~Alexander~Leonessa
\thanks{Authors are with the Department
of Mechanical Engineering, Virginia Polytechnic Institute and State University, Blacksburg,
VA, 24060 USA }
\thanks{* Corresponding Author (cwh@vt.edu)}
\thanks{Manuscript received April 19, 2005; revised August 26, 2015.}}

%
%

\markboth{Journal of \LaTeX\ Class Files,~Vol.~14, No.~8, August~2015}%
{Shell \MakeLowercase{\textit{et al.}}: Bare Demo of IEEEtran.cls for IEEE Journals}
%



\maketitle

\begin{abstract}
In this work, the joint-control strategy is presented for the humanoid robot, PANDORA, whose structural components are designed to be compliant. As opposed to contemporary approaches which design the elasticity internal to the actuator housing, PANDORA's structural components are designed to be compliant under load or, in other words, \textit{structurally elastic}. To maintain the rapid design benefit of additive manufacturing, this joint control strategy employs a disturbance observer (DOB) modeled from an ideal elastic actuator. This robust controller treats the model variation from the structurally elastic components as a disturbance and eliminates the need for system identification of the 3D printed parts. This enables mechanical design engineers to iterate on the 3D printed linkages without requiring consistent tuning from the joint controller. Two sets of hardware results are presented for validating the controller. The first set of results are conducted on an ideal elastic actuator testbed that drives an unmodeled, 1 DoF weighted pendulum with a 10 kg mass. The results support the claim that the DOB can handle significant model variation. The second set of results is from a robust balancing experiment conducted on the 12 DoF lower body of PANDORA. The robot maintains balance while an operator applies 50 N pushes to the pelvis, where the actuator tracking results are presented for the left leg. 

\end{abstract}


%
\IEEEpeerreviewmaketitle

\section{Introduction}
Recently, a labor crisis has affected the global market where countries such as the U.S., Germany, South Korea, and China  
need upwards of 5.5\% of their population to fulfill their current business demand \cite{liu2024why}. According to the World Health Organization, the healthcare industry is already highly in demand where, in 2021, global spending rose to 9.8 trillion USD, or 10.3\% of the world's GDP \cite{world2024global}. As the global age is continuing to rise, these costs and healthcare worker demands are only expected to increase \cite{uscb2023how, silvera2024robotics, ellingrud2023generative}. In addition, demands on the manufacturing industry have also risen where, according to the U.S. Chamber of Commerce, 622,000 jobs have yet to be filled as of January 2024 \cite{ferguson2024understanding}. The current labor crisis and increases to global costs and demand drive the need for technological innovation \cite{ellingrud2023generative}. 


To overcome this increasing demand, artificial intelligence (AI) and robotics can tackle a wide range of tasks for achieving efficiency improvements \cite{nam2021adoption, gibney2024ai}. AI can leverage several types and large quantities of data to identify patterns, build models, and generate images, video, and audio \cite{ellingrud2023generative}. Robots, on the other hand, can embody this intelligence to physically affect their surroundings, capable of reliably completing routine behaviors quickly and autonomously \cite{gibney2024ai}. In the U.S. automation industry alone, it is estimated that the combination of AI and robotics will take over 29.5\% of current tasks by 2030 \cite{ellingrud2023generative}. 

In order to meet these growing expectations, several challenges affect the acceptance of AI and robots. These challenges include distrust, poor understanding, lack of convenient interfaces, safety concerns, and general difficulty with scaling to real-world applications \cite{silvera2024robotics, nam2021adoption, gibney2024ai}. Humanoid robots are uniquely positioned to address many of these challenges since their human-centric form factor allows them to ergonomically transition to the existing workplace. Companies such as Mercedes, BWM, Hyundai, Volkswagen, Amazon, Toyota, and Tesla have created collaborations or are developing their own humanoid robots for conducting research and commercialization \cite{weatherbed2024mercedes, staff2024bmw, hyundai2024hyundai, wu2024humanoid, amazon2022introducing, toyota2024boston, reuters2024tesla}. Humanoids offer potential in dealing with repetitive and otherwise boring tasks, while also extending human-like dexterity to dangerous environments through teleoperation \cite{dafarra2024icub3}. Humanoids are already being developed for applications such as construction and automation \cite{staff2024bmw, weatherbed2024mercedes, kheddar2019humanoid, reuters2024tesla, boston2024sick}, improvised explosive device (IED) response \cite{jorgensen2019deploying}, entertainment \cite{dafarra2024icub3, boston2024do, johnson2016exploring}, healthcare \cite{porta2022towards, abhishek2021humanoid}, space exploration \cite{diftler2011robonaut, wimbock2009experimental, radford2015valkyrie}, and even caring for children \cite{robins2004effects, kozima2004humanoid, wood2021developing, simo2006humanoid, dautenhahn2009kaspar}.

Humanoids are often designed to be bipedal which, unlike other robots, requires sophisticated technology to maintain balance during locomotion. State-of-the-art hierarchical controllers require a multitude of sensors for measuring the robot state including absolute and incremental encoders, linear and multi-axis load cells (FT sensors), inertial measurement units (IMU), accelerometers, pressure sensors, LiDAR, stereo cameras, and even temperature sensors \cite{knabe2015design, asfour2018armar, herron2024pandora, englsberger2014overview, penco2024mixed, dafarra2024icub3}. In addition, the mechanical design is often quite complicated since these robots encounter dynamic loading scenarios from impacts during walking. For TREC's previous humanoids, THOR and ESCHER \cite{knabe2015design}, the robots contained more than 100 highly intricate metal parts to achieve desired weight and strength metrics. All of the components were made via subtractive manufacturing (SM) from materials such as steel, titanium, and aluminum. Due to design complexity, nearly all of parts were multi-sided and required a expert machinist to achieve the tolerance expectations for assembly and operation \cite{fuge2023design}. In fact, a majority of the current humanoid robots can be expected to utilize SM since the costs are reduced at higher volumes \cite{sathish2022comparative}.

While SM is more popular, additive manufacturing (AM) offers significant advantages in terms of design capabilities. Whereas SM cuts away at a block of raw material, AM adds material layer-by-layer to produce the desired component \cite{sathish2022comparative}. AM is resource efficient and lowers the costs for prototyping, eliminates the need for complex fixtures and tools, lowers the requirement on operator expertise, and is quite accessible \cite{bhatia2023additive, sathish2022comparative, jayawardane2023sustainability}. AM also allows the designer to create highly intricate parts that otherwise must be split into several smaller components or are impossible to manufacture though SM. AM supports a wide range of materials including metals, polymers, ceramics, resins, wood, powders, glass, and even smart materials. This large set of options make AM applicable to virtually every industrial sector \cite{bhatia2023additive, srivastava2022review}. Robotics utilizes 3D printing for a variety of applications, but typically incur significant drawbacks from dimensional accuracy, part sustainability, and overall reliability \cite{sathish2022comparative, jayawardane2023sustainability}. 

Despite these difficulties, AM has already made a significant impact on the humanoid robotics field. The hydraulic Atlas humanoid robot from Boston Dynamics utilized metal 3D-printing to manufacture several high-strength linkages using a combination of aluminum and titanium \cite{boston2024atlas}. This technology enabled engineers to embed hose-less routing of the hydraulic fluid directly into the structural components which otherwise would have been impossible to manufacture via SM \cite{alright2016boston}. Overall, this approach simplified the design, lowered part and interface count, and allowed Atlas to execute highly agile maneuvers \cite{alright2016boston, boston2024sick, boston2024do}. The NimbRo-OP2X is an open-sourced mid-size humanoid robot that utilized 3D printing for the structural components to maintain a low weight of 19 kg. The lower weight allowed for more torque to be utilized for faster behaviors and enabled the robot to win the RoboCup in 2018 \cite{ficht2018nimbro}. Finally, iCub is a child sized robot designed to study the application of humanoids in disaster assistance, ergonomics, automation, child care, and even teleoperation \cite{metta2010icub, dafarra2024icub3}. iCub's 3D printed tactile skin measures the pressure applied to the full body and provides a robust form of feedback with regards to how an individual can interact with the robot \cite{maiolino2013flexible, forestier2024biomimetic}. In the case of teleoperation, the operator wears a suit that provides haptic feedback at the body location where forces are being applied to iCub and results in a more immersive experience \cite{dafarra2024icub3}. In this work, 3D printing is utilized to embed the material compliance into the structural linkages to take the place of elastic components typically designed to sit within the actuator housings.

In the previously mentioned works, 3D printing is used for creating complex components that may not be manufacturable through SM which leads to lower part count and simpler designs. Alternatively, this work leverages the material choice to incorporate compliance directly into the structural components, eliminating the need for additional interfaces, bolts, and a dedicated elastic component. This design choice introduces controls and estimation challenges that, to the authors knowledge, have not been uncovered in prior works for humanoid robots. As discussed in \cite{herron2024pandora}, PANDORA has a hybrid design where a majority of the components are manufactured using 3D printed PLA+ with SM utilized for high-stress metal components. The choice of material for the 3D printed parts has introduced significant compliance, removing the need for elastic components within the actuators. In this work, the authors use the term \textit{structural elasticity} to imply that the main structural components are designed to be compliant rather than including additional components within the actuator housing. In prior works, humanoid robots often leverage elasticity to avoid damaging components from high magnitude impacts experienced during walking where the design and modeling of the elastic component is critical for tracking joint trajectories \cite{paine2013design, hopkins2015embedded, kong2011compact, pratt1995series, pratt2004series}. However, heavy modeling of the elastic component is difficult for PANDORA and adds additional constraints which eliminates the advantage of rapid design iterations and manufacturing that AM offers. 

In order to maintain these AM advantages without compromising the control performance, this work utilizes a DOB-based approach to handle model nonlinearities such as friction and unknown dynamics. This approach avoids difficult modeling of structural components that vary based on a variety of AM factors such as material, print orientation and speed, bed temperature, wall thickness, support structure, layer height, and infill pattern \cite{herron2024pandora}. As a result, this robust control choice allows the mechanical engineers to constantly iterate on mechanical structure while still achieving joint control objectives. This work provides a general joint-space control approach for humanoid robots which exhibit structurally elastic dynamics that are unmodeled. In addition, this work discusses key details for joint control that are critically important for replication, but are often disregarded from publications. This design approach can be leveraged beyond humanoid robots for a variety of applications that require compliance including human-robot interactions (HRI), exoskeletons, and soft robotics. The main contributions of this work are
\begin{itemize}
    \item An overview of the joint control approach for a \textit{structurally elastic} humanoid robot where the elastic linkages do not require system identification\\
    \vspace{-0.2835 cm}
    \item Key details for replicating humanoid robot control from the output of the Whole-Body Controller (WBC) to the motor commands. These details include leaky integration, efficient digital software design for representing continuous systems, and kinematic estimation challenges. \\
    \vspace{-0.2835 cm}
    \item Two sets of results which validate the joint control approach for handling significant model variation. The first is conducted on an elastic actuator testbed driving an unmodelled, 1 DoF, 10 kg pendulum and demonstrates how the DOB can handle significant model variation. The second is a balancing experiment on PANDORA's 12 DoF lower body while an operator applied 4 disturbances to the pelvis. The joint control algorithm from this work is utilized for controlling the 12 DoF joints despite the linkage compliance being elastic and unmodelled.
\end{itemize}

The rest of the paper is organized as follows: Section \ref{sec:structurally_elastic_humanoid} defines the term \textit{structural elasticity} and compares PANDORA's elasticity approach to contemporary elastic actuation methods of other legged robots, Section \ref{sec:humanoid_control_overview} describes PANDORA's control hierarchy and mechatronics design, Section \ref{sec:joint-space-control} discusses the joint-space controller design including structural elasticity modeling, leaky integration, impedance, and efficient software implementation, Section \ref{sec:results_and_discussion} presents two sets of results for validating the joint-space control approach, and Section \ref{sec:conclusion} concludes the paper.

\newpage

\section{Structurally Elastic Humanoid} \label{sec:structurally_elastic_humanoid}

\begin{figure}[!t]
\centering
\includegraphics[width=\linewidth]{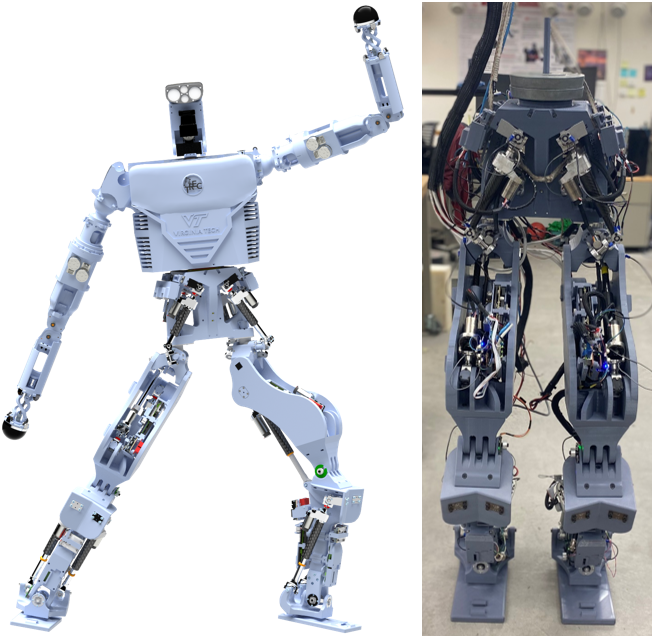}
\caption{Full rendering of PANDORA humanoid robot (left) with the lower-body fully built and operational (right)}
\label{fig:pandora-robot}
\vspace{-0.5 cm}
\end{figure}

As displayed in Fig. \ref{fig:pandora-robot}, PANDORA is 1.9 m and weighs 49 kg, with 12 degrees of freedom (DoF) in the lower body and 20 in the upper body \cite{fuge2024innovative}. PANDORA's lower-body is fully built and driven using linear actuators repurposed from THOR and ESCHER \cite{knabe2015design}. The THOR and ESCHER humanoid robots were designed from aluminum, steel, and titanium components whereas PANDORA features a hybrid design composed of aluminum and 3D printed components made from PLA+. Utilizing 3D printing for the structural components significantly reduced the part count by over 50\% compared to her predecessors and lowered the assembly time to only 8 hours with 2 people \cite{herron2024pandora, fuge2023design}. PANDORA's lower body is composed of 1 and 2 DoF joints where the hip yaw/roll and ankle pitch/roll are simultaneously controlled by sets of 2 actuators and the hip and knee pitch are individually controlled by single actuators.

\begin{figure}[!t]
\centering
\includegraphics[width=\linewidth]{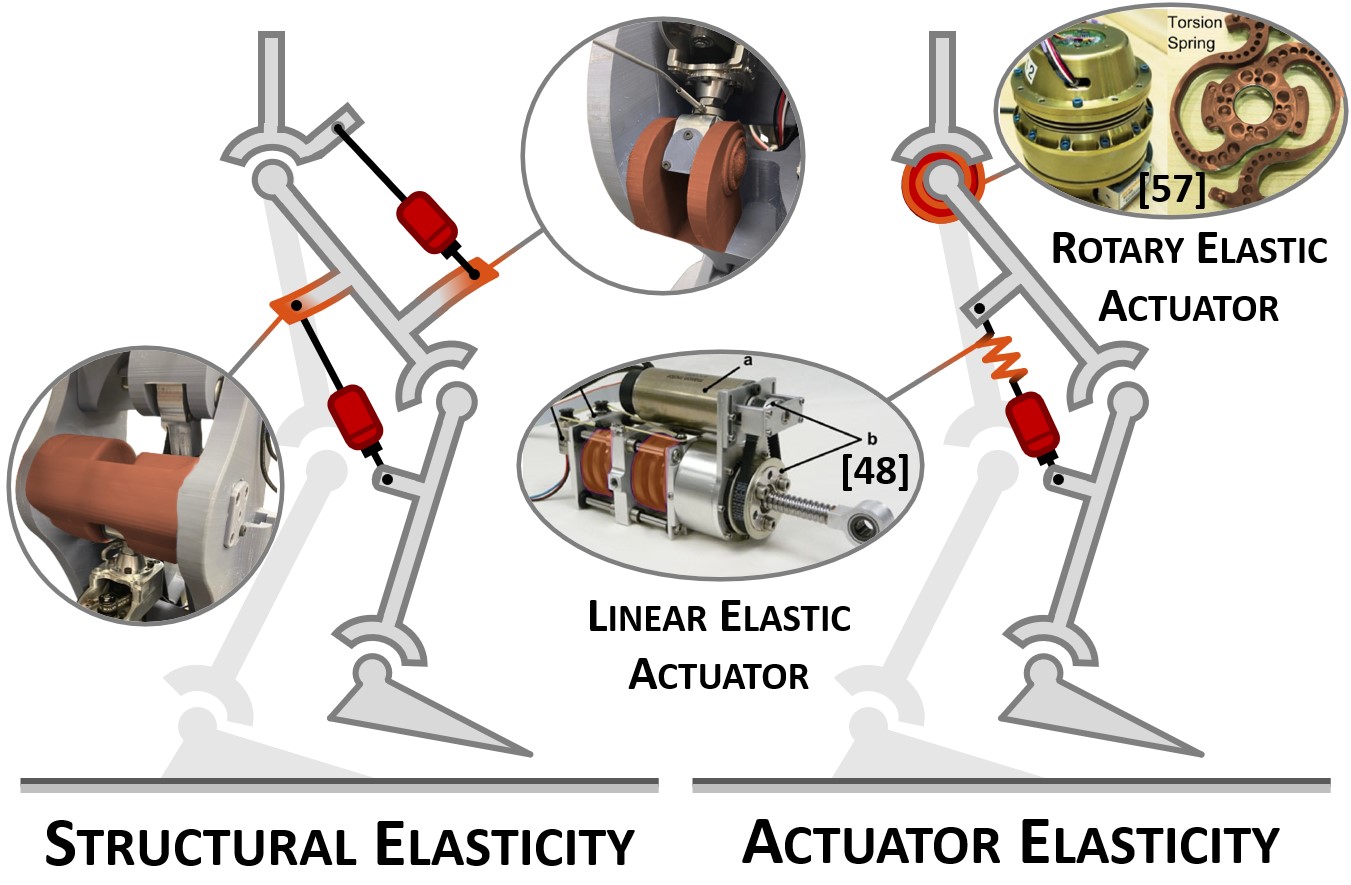}
\caption{Comparison between structural elasticity from PANDORA (left) and conventional elasticity approaches (right) within the actuators.}
\label{fig:elasticity-comparisons}
\vspace{-0.46 cm}
\end{figure}

For THOR and ESCHER, the joints were driven via linear elastic actuators where a titanium bar acts as the compliant mechanism \cite{knabe2015design, hopkins2015embedded}. For PANDORA, this compliant mechanism has been removed and the linkage is specifically designed to provide compliance. This work uses the term \textit{structurally elastic} to capture the idea that portions of the linkages are designed to be compliant under load. As displayed in Fig. \ref{fig:elasticity-comparisons}, the actuator mounting points have been carefully designed through static loading of individual linkages and dynamic loading of the full robot during balancing. However, a considerable amount of research is still necessary for developing models and estimating the expected performance of loaded 3D printed components \cite{herron2024pandora}. Future work will discuss this design approach, focusing on design heuristics for AM of larger robotic systems. In general, the printing material, print settings, and part design can have a considerable impact on the robot's performance especially in state estimation and control. This work presents a joint control strategy which does not require rigorous modeling of the 3D printed components and can be generalized to a variety of elastic robotic systems.

Actuator elasticity provides protection from shock impulse, increased force-to-torque stability margins, and efficiency improvements \cite{pratt1995series, pratt2004series}. Legged robots such as quadrupeds and humanoid robots often benefit from elastic mechanisms to protect the hardware from high force impacts spikes during walking. This approach has been featured on Valkyrie \cite{paine2015actuator,paine2013design}, THOR , ESCHER \cite{knabe2015design}, UXA-90 \cite{lanh2023hybrid, truong2020design}, iCub \cite{tisi2017design, nava2017momentum}, and WALK-MAN \cite{negrello2016walk}. As displayed in Fig. \ref{fig:elasticity-comparisons}, these actuators are designed with either linear or rotary motion to drive parallel or in-line with the joint, respectively. A considerable amount of effort goes into the design of elastic actuators to achieve desired properties such as the power output, size and weight, efficiency, back-drivability, backlash, and stiffness. These design parameters can be achieved through careful selection of the transmission mechanisms such as gears and belts, ball or lead screws, motors, and spring types \cite{paine2013design, kong2011compact}. The spring mechanism can be designed to provide a fixed or configurable stiffness where a configurable stiffness allows the passive energy storage device to be modifiable online \cite{orekhov2013configurable, paine2013design}. This design approach can be advantageous for legged robots where prior research suggests that humans and other animals vary their leg stiffness for efficient hopping and running behaviors \cite{alexander1990three, ferris1998running}. 

Humanoid robots are typically designed such that the linkages are stiff using metal materials such as aluminum, steel, and titanium. This design approach simplifies the robot model where the kinematics can be calculated using the measured joint angles and estimated base configuration. Obtaining an accurate kinematics model is important for estimating the footstep locations and center of mass (CoM) which are critical for robot control. These components can be difficult to estimate even for stiff robots, where linkages can bend under load \cite{koolen2016design}. PANDORA's linkages also exhibit this behavior where the footstep position estimation can be inaccurate up to 5 cm during robust balancing \cite{herron2024pandora}. As discussed in the following section, this challenge can affect the performance at each level of the hierarchical controller.



\begin{figure*}[!t]
\centering
\includegraphics[width=\textwidth]{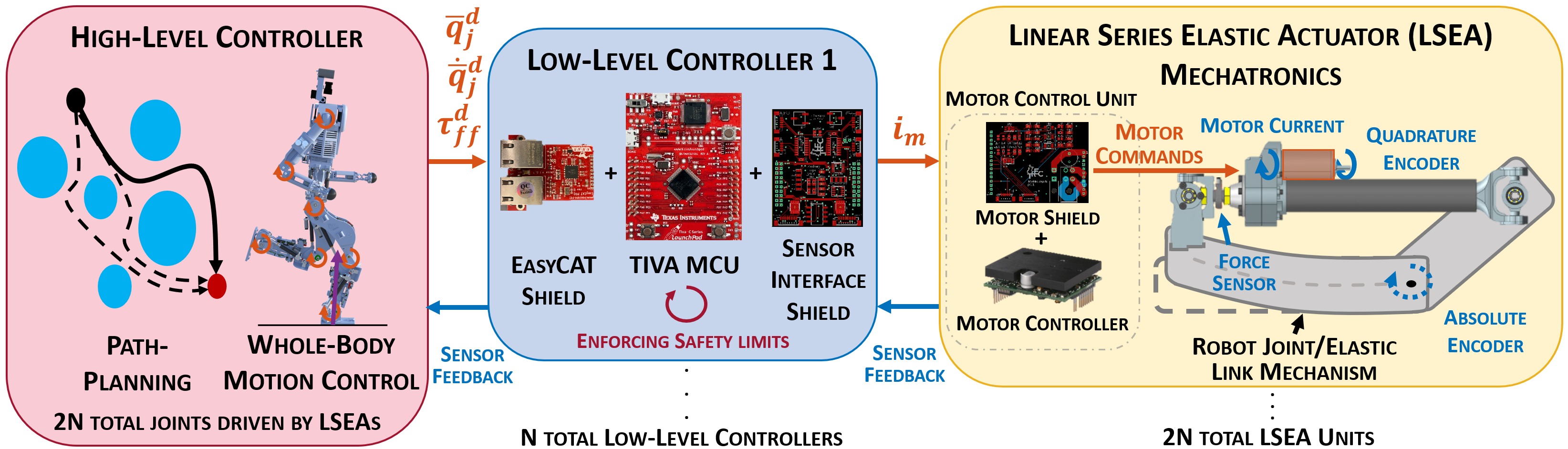}
\caption{PANDORA's control hierarchy is broken down into high and low-level controllers utilizing motor current, quadrature and absolute encoder, and force sensor feedback from an actuator/joint pair.}
\label{fig:humanoid-control-overview}
\vspace{-0.5 cm}
\end{figure*}
\newpage

\section{Humanoid Control Overview} \label{sec:humanoid_control_overview}

This section focuses on PANDORA's control hierarchy broken into high and low levels of control and discusses PANDORA's mechatronics design including the layers of sensing and feedback between the control layers. This control and mechatronics approach is consistent with other legged robots that utilize similar planning approaches and feedback sensor types \cite{knabe2015design, paine2015actuator, dafarra2024icub3, radford2015valkyrie, englsberger2014overview}. This paper focuses on presenting the joint-level control approach which is designed within the Low-Level Controller (LLC).

The high-level controller is composed of a planner and Whole-Body Controller (WBC). The planner is typically a combination of path, footstep, and CoM planners that often utilize a reduced-order model. The path planner decides the route to take attempting to avoid any obstacles within the area to reach the objective. This route can be used to develop a feasible footstep plan and subsequently obtain the CoM motion for the WBC
The CoM planner can be composed in a variety of ways using methods such as Model Predictive Control (MPC) \cite{di2018dynamic}, Linear Inverted Pendulum (LIP) dynamics \cite{radford2015valkyrie}, and the Divergent Component of Motion (DCM) \cite{hopkins2015dynamic, englsberger2014overview}. Some planners do not require precise footstep locations and rely on well tuned heuristics \cite{raibert1986legged, di2018dynamic}, but this can lead to stability challenges if terrains are too rough \cite{wiedebach2016walking}. PANDORA utilizes the DCM approach which indirectly stabilizes the CoM based on the DCM coordinate definition \cite{hopkins2015compliant}. Therefore, the DCM coordinate can be driven from footstep to footstep to result in feasible CoM and GR force trajectories. Recent work has extended beyond linear motion to include an orientation objective \cite{herron2024angular}.
The CoM trajectories are provided as a centroidal momentum objective for the WBC \cite{koolen2016design}. The WBC is a optimization problem where the cost function is composed of motion tasks and often designed to be convex to achieve real time constraints. The motion tasks can be designed from desired momentum rates of change, spatial and joint accelerations, and position biases that orchestrate a specific desired behavior such as balancing, walking, and running. This optimization problem considers the full order dynamics, as well as, position and torque feasibility constraints. The output of the WBC are joint accelerations, $\ddot{q}_{\rm j}^{\rm d}$, and ground reaction wrenches that can be used to calculate joint torques, $\ddot{\tau}_{\rm ff}^{\rm d}$, via inverse dynamics \cite{koolen2016design, hopkins2015compliant}. For PANDORA, the High-Level Controller is designed using the IHMC Open Robotics Software (ORS) \cite{ihmcORS} and TREC Robotics Software (TRS) \cite{trecTRS} within a high-powered PC.

As discussed in Section \ref{sec:joint-space-control}, the desired joint accelerations,  $\ddot{q}_{\rm j}^{\rm d}$, can be used to identify the desired joint positions, $\bar{q}_{\rm j}^{\rm d}$ and velocities, $\dot{\bar{q}}_{\rm j}^{\rm d}$, through leaky integration. The leaky integration simply biases these trajectories to avoid instabilities that result from integrator windup \cite{jorgensen2020towards}. The leaky integration step is completed in the High-level Controller whereas the remainder of the joint-level controller is performed on the LLCs. The desired joint trajectories, $\bar{q}_{\rm j}^{\rm d}$, $\dot{\bar{q}}_{\rm j}^{\rm d}$, and $\ddot{\tau}_{\rm ff}^{\rm d}$, are sent from the High-Level Controller to the each of the LLCs at 500 Hz. The LLCs are hardware components that are responsible for collecting and conditioning sensor feedback, computing motor commands, checking and enforcing safety limits, and communicating with the high-level system \cite{herron2023design}. Each LLC can collect the sensor feedback and control the actuators of 2 pairs of joints/actuators. PANDORA's lower body is designed such that the 1 and 2 DoF joint configurations are driven using linear actuators and are neatly paired for joint control. 

As displayed in Fig. \ref{fig:humanoid-control-overview}, the sensor feedback for the Linear Series Elastic Actuator (LSEA) mechanism is collected by an LLC and sent back to the High-Level Controller for the following planning and WBC computation. As discussed in Section \ref{sec:structural-elasticity}, PANDORA's structural components are designed to be partially compliant and provides useful elasticity for the actuators. Each LLC can measure 2 sets of sensors from a joint/actuator pair where several forms of sensor feedback are useful for interpretting the robot state.
A quadrature and absolute encoder are utilized for measuring position of the motor and joint, respectively. The motor current of the actuator is conditioned using a notch filter to attenuate the power supply mains hum and passive low-pass RC filter to attenuate motor noise and measured via the LLC's ADC. The applied actuator force is measured from a load cell with a 2nd order, unity-gain, Sallen-Key analog filter to attenuate high frequency noise.
The actuator is controlled by sending motor current commands, $i_m$, via pulse width modulation (PWM), from the LLC to a motor control unit (MCU) at 20 kHz. The MCU is an off-the-shelf component that utilizes the motor's hall-sensor feedback to achieve $i_m$ \cite{herron2023design}. These $i_m$ commands are determined from the joint-space controller discussed in the following section.

\begin{figure*}[!t]
\centering
\includegraphics[width=\linewidth]{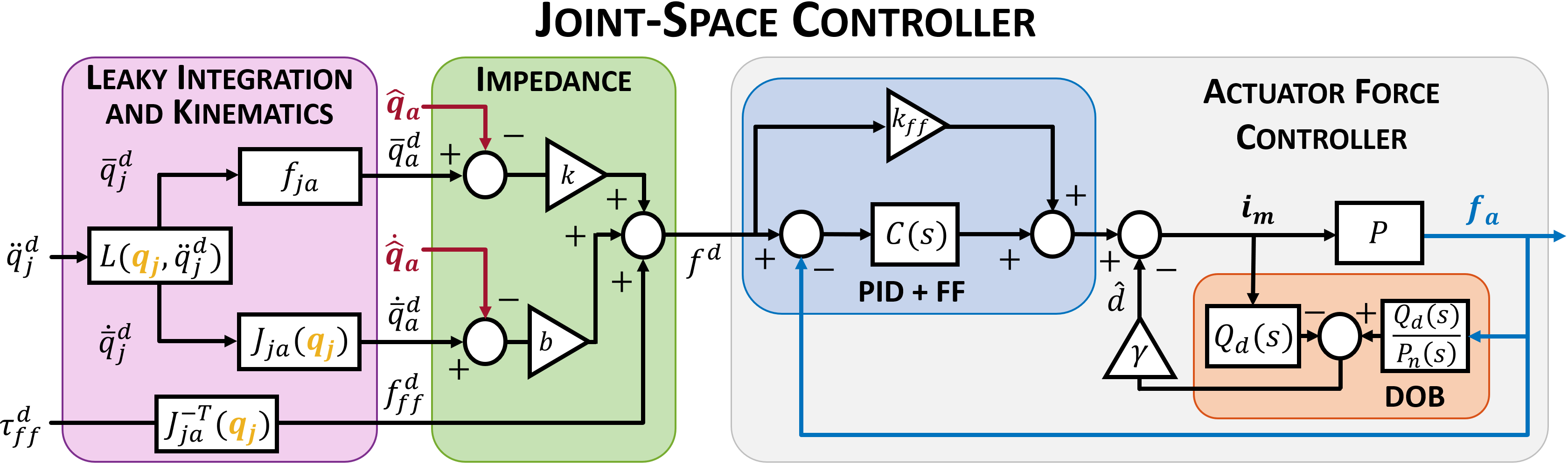}
\caption{Overview of the Joint-Space Controller that uses leaky integration and kinematics to identify actuator objectives and impedance and force Control to achieve those trajectories.}
\label{fig:joint-space-controller}
\vspace{-0.5 cm}
\end{figure*}
\newpage

\begin{figure}[!t]
\centering
\vspace{-0.01 cm}
\includegraphics[width=\linewidth]{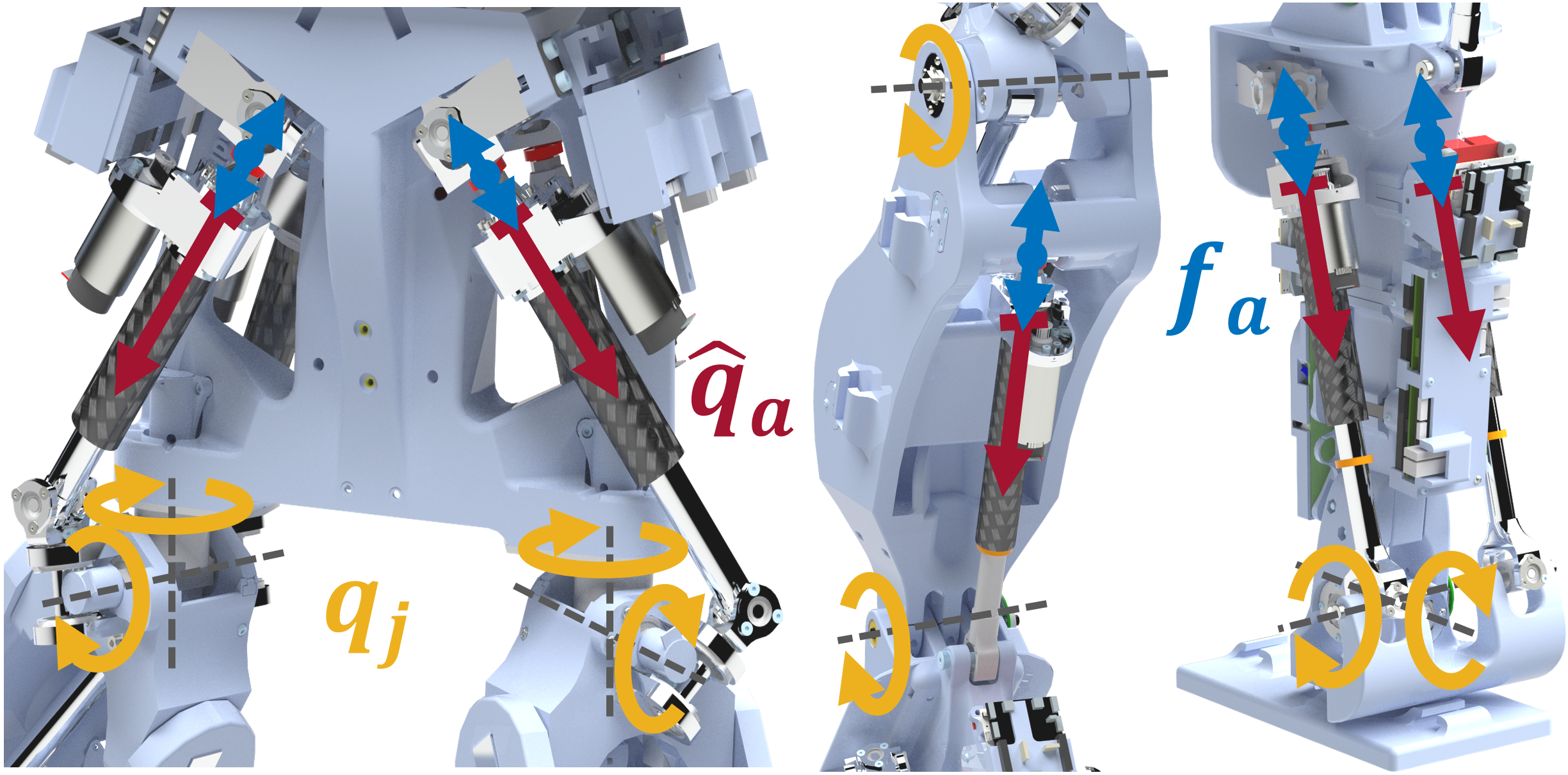}
\caption{Overview of lower body measurement axes. These measurements relate to the kinematic transform from the joint and actuator spaces.}
\label{fig:lower-body-measurements}
\vspace{-0.5 cm}
\end{figure}

\section{Joint-Space Control} \label{sec:joint-space-control}
This section describes the entire joint-space controller from the WBC joint trajectories to the actuator control commands where the full diagram is displayed in Fig. \ref{fig:joint-space-controller}.
As discussed in \cite{herron2024pandora, fuge2023design}, PANDORA is composed of 1 and 2 DoF joints driven by linear actuators in parallel. This  approach is particularly valuable for highly condensed joints and allows for two actuators to contribute torque in the same direction. As seen in Fig. \ref{fig:lower-body-measurements}, the hip yaw/roll and ankle pitch/roll joints act as 2 DoF systems where two actuators apply torque to both joints simultaneously. The knee and hip pitch are 1 DoF joints driven by a single actuator in parallel. This difference in joint mechanisms only affects the kinematics conversion, where the desired joint acceleration and torque trajectories are mapped into desired actuator position, velocity, and force trajectories. To provide additional stability, these actuator trajectories are composed into a single desired force trajectory using virtual impedance (spring and damper) gains. 
\begin{table}[]
    \centering
    \resizebox{\linewidth}{!}{%
    \begin{tabular}{|c|c|c|c|c|c|c|c|c|c|}
        \hline
        \rowcolor{lightgray}
        & \textbf{Actuator} & $\pmb k$ & $\pmb b$ & $\pmb k_{\rm ff}$ & $\pmb k_{\rm p}$ & $\pmb k_{\rm i}$ & $\pmb k_{\rm d}$ & $\pmb \lambda_{\rm c}$ & $\pmb \gamma$ \\
        \hline
        \multirow{6}{3 pt}{\begin{sideways}\textbf{Left}\end{sideways}}& F. Hip & 40 & 5 & 3.2 & 15 & 4 & 2.5 & 3.5 & 0.8 \\
        \cline{2-10}
        & B. Hip  & 40 & 5 & 3.2 & 15 & 4 & 2.5 & 3.5 & 0.8 \\
        \cline{2-10}
        & Thigh & 30 & 5 & 3.2 & 20 & 5 & 2.5 & 3.5 & 0.4 \\
        \cline{2-10}
        & Knee & 30 & 5 & 3.2 & 20 & 5 & 2.5 & 3.5 & 0.5 \\
        \cline{2-10}
        & L. Ankle & 40 & 2.5 & 3.2 & 15 & 4 & 2.5 & 3.5 & 1.0 \\
        \cline{2-10}
        & R. Ankle & 40 & 2.5 & 3.2 & 15 & 4 & 2.5 & 3.5 & 1.0 \\
        \hline
        \multirow{6}{3 pt}{\rotatebox[origin=c]{90}{\textbf{Right}}}& F. Hip & 40 & 5 & 3.2 & 15 & 4 & 2.5 & 3.5 & 0.8 \\
        \cline{2-10}
        & B. Hip & 40 & 5 & 3.2 & 15 & 4 & 2.5 & 3.5 & 0.8 \\
        \cline{2-10}
        & Thigh & 30 & 5 & 3.2 & 20 & 5 & 2.5 & 3.5 & 0.4 \\
        \cline{2-10}
        & Knee & 30 & 5 & 3.2 & 20 & 5 & 2.5 & 3.5 & 0.5 \\
        \cline{2-10}
        & L. Ankle & 40 & 2.5 & 3.2 & 15 & 4 & 2.5 & 3.5 & 1.0 \\
        \cline{2-10}
        & R. Ankle & 40 & 2.5 & 3.2 & 15 & 4 & 2.5 & 3.5 & 1.0 \\
        \hline
    \end{tabular}}
    \captionof{table}{Gain settings for controlling PANDORA's lower body actuators. The impedance gains, $k$ and $b$, have units of $\rm{N}/\rm{m}$ and $\rm{N\,s}/\rm{m}$, respectively.}
    \label{tab:joint-controller-gains}
\vspace{-0.6 cm}
\end{table}
Finally, in order to achieve force control, a PID + Feedforward (FF) controller is utilized alongside a Disturbance Observer (DOB). The DOB is a critical component for dealing with nonlinearities in the system such as stiction, backlash, and model uncertainties. In this way, the DOB is responsible for dealing with the structural elasticity by providing feedback to maintain a nominal plant model such that the controller can remain effective, regardless of the actuator's placement on the robot. The following subsections will explain the main components of the joint control strategy in more detail where all of the various gains for each of the actuators in the lower body can be seen in Table \ref{tab:joint-controller-gains}. 

\subsection{Structural Elasticity Modeling} \label{sec:structural-elasticity}
Early on in PANDORA's development, numerous materials were tested including PLA, PLA+, ULTEM, Nylon, ABS, and ABS with embedded carbon fiber where each provided a different stiffness \cite{herron2024pandora, fuge2023design, fuge2024innovative}. The authors found it difficult to model or identify the elasticity of the various linkages due the high number of variables that impact 3D printed component performance including material, part orientation, printer quality, print time, bed temperature, layer and wall thickness, support and infill pattern, and layer height \cite{herron2024pandora, fuge2024innovative}.
In addition, \cite{fuge2023design, herron2023design, fuge2024innovative} discusses deficiencies in modeling and simulating 3D printed materials, where empirical testing is still required to identify parameters such as stiffness and strength. 
Therefore, the authors required a robust control strategy which allowed the mechanical design engineers the ability to consistently iterate on the structural components without affecting the joint control performance. 

\begin{figure}[!t]
\centering
\includegraphics[width=\linewidth]{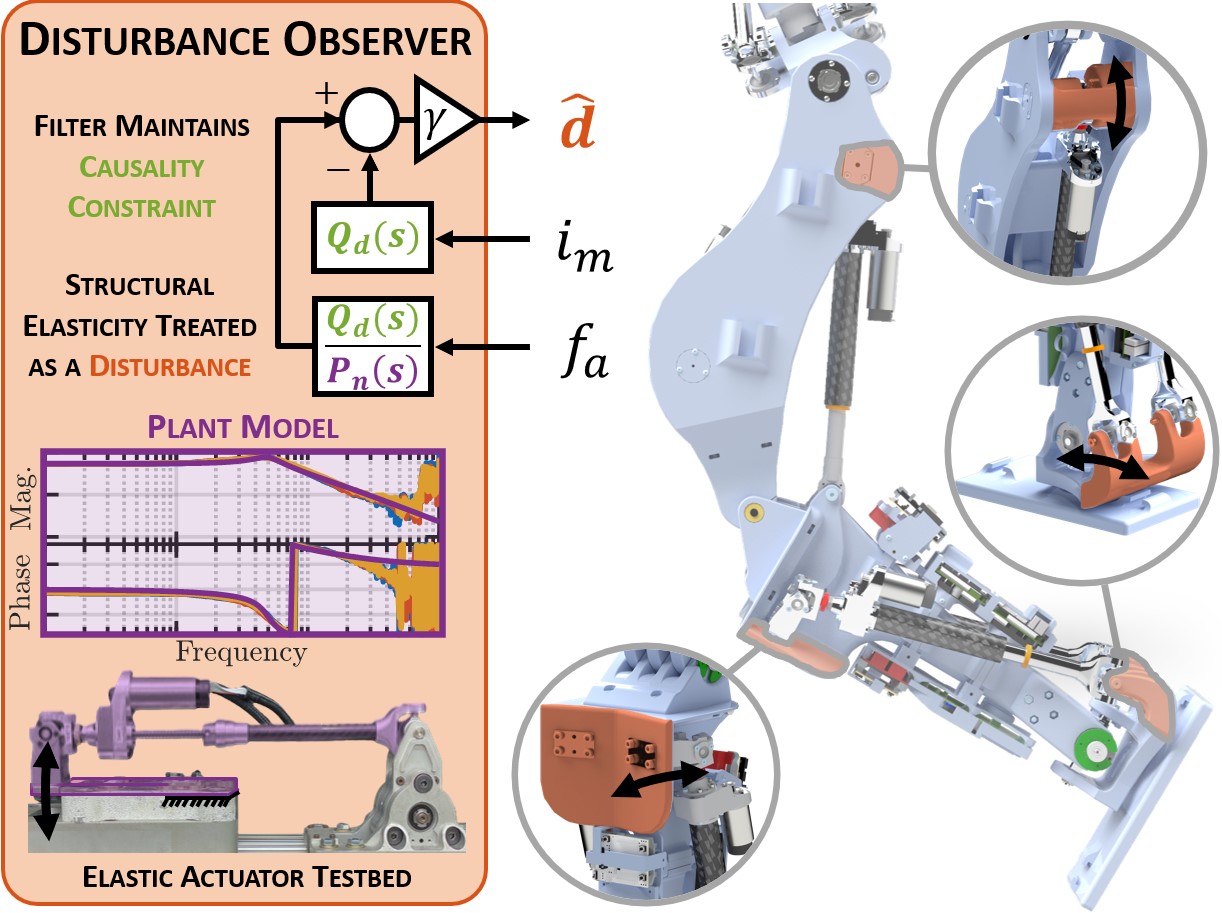}
\caption{The DOB plant model is designed using an elastic actuator test bed and compensates for the structural elastic components by providing a disturbance as feedback in the actuator force control loop.}
\label{fig:dob-design}
\vspace{-0.5 cm}
\end{figure}

One of the most popular approaches for controlling humanoid robots is utilizing linear actuators attached to drive the joint in parallel. An elastic component is often connected to the gear train acting as a passive spring resulting in a mechanical low-pass filter between the motor and its loads \cite{pratt1995series, hopkins2015embedded, paine2013design, oh2016high, yildirim2021design, aguirre2020lower, paine2015actuator}. For a linear system, this combination is known as an LSEA and is often paired with the DOB control strategy. As discussed in \cite{hopkins2015embedded, paine2013design, oh2016high, aguirre2020lower, haninger2020safe, paine2015actuator}, DOBs are designed to provide feedback to a system based on a nominal plant model found through empirical testing. The DOB can have different configurations and shares similar properties to adaptive time-delay controllers \cite{jin2016model} As displayed in Fig. \ref{fig:joint-space-controller} and \ref{fig:dob-design}, the DOB compares the previous commanded input to the current measured output and provides this as a disturbance feedback to the actuator force control loop. This approach has been known to handle nonlinearities such as stiction, backlash, and additional unmodeled loads and dynamics \cite{hopkins2015embedded, paine2013design, haninger2020safe}.

For previous generation TREC humanoids, THOR and ESCHER \cite{knabe2015design}, LSEAs were designed with a titanium beam acting as its elastic element. These titanium beams were removed for PANDORA, where the 3D printed structurally elastic linkages are designed to provide compliance. Early on in the DOB design phase, the authors removed the titatium beam from the elastic actuator testbed and replaced it with a 3D printed mount expecting to obtain an input/output model that sufficiently matched the elastic behavior of PANDORA's linkages. However, the mount was considerably stiffer than the link dynamics and resulted in a high (+7) order model. In an attempt to avoid the challenge of designing a mount which mimicked the structurally elastic linkages, the DOB plant model, $P_n(s)$, is designed based on the elastic actuator testbed with a titanium bar because this system provided excellent control for THOR and ESCHER and is low order \cite{hopkins2015embedded}. 

\begin{figure}[!t]
\centering
\includegraphics[width=\linewidth]{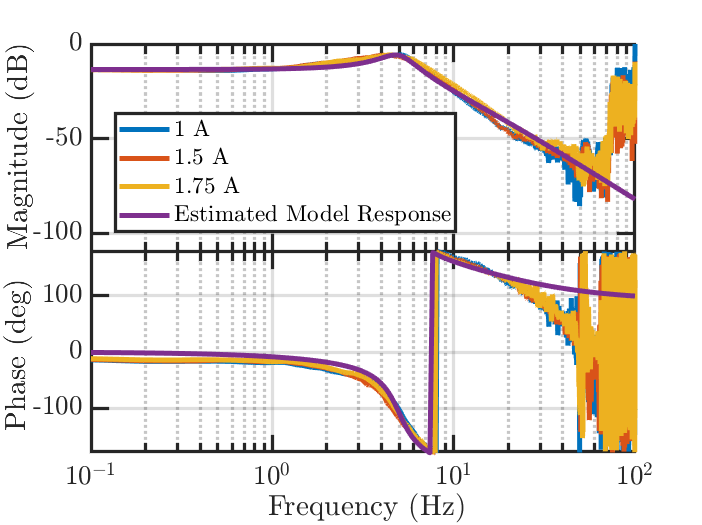}
\caption{Open-loop bode plot of the elastic actuator testbed at various input amperage amplitudes. The input of the system is the input motor current where the output is the measured force from actuator's linear load cell.}
\label{fig:ol-plant-bode}
\vspace{-0.5 cm}
\end{figure}

An overview for this approach is displayed in Fig. \ref{fig:dob-design}, including an image of the elastic actuator testbed (purple) and various regions of the linkages which are designed to be compliant (orange). 
For building the DOB, the nominal plant model, $P_n(s)$, was identified using the elastic actuator testbed where the titanium beam acts as a compliant mechanism. This SISO model has an input motor current, $i_m$, and an output force, $f_o$, where each of these signals can be measured via the LLC \cite{herron2023design}. 
For obtaining the model, $i_m$ is designed as an exponential chirp signal and sent to the actuator in open-loop (OL) simply to measure $f_o$ for an input/output comparison. During the experiment, the testbed is locked in place such that only the LSEA's compliant titanium beam provides motion. As displayed in Fig. \ref{fig:ol-plant-bode}, the plant model's OL bode plot is determined for a frequency range of 0.1 - 100 Hz where the system shows a similar response for an input amperage of $1$, $1.5$, and $1.75$ $\rm{A}$. Each experiment took 2 minutes to conduct where the exponential chirp signal is generated using the algorithm discussed in \cite{herron2024software}.
Using this data, the nominal plant model, $P_n(s)$, was generated using the System Identification Toolbox in {\tt MATLAB}. Thus, the estimated nominal plant model is represented in the Laplace domain as
\begin{equation} \label{eqn:nominal-plant}
    P_n(s) = \frac{208.8}{0.01s^3 + 1.13s^2 + 23.04s + 987.0}.
\end{equation}
Note the similarity between the nominal plant model and the system response in Fig. \ref{fig:ol-plant-bode}, where the model generally captures the system motion up to 30 Hz. Starting around 20 Hz, the motor housing begins to induce vibration which creates to a drop in phase, an increase in noise magnitude, and begins to exhibit model deviation from the physical responses. This critical dynamic effect was identified by removing the belt which connects the motor to the linear actuator's gear train and generating a bode plot from the same OL chirp signal. An important detail regarding (\ref{eqn:nominal-plant}) is that the system is unmistakably 3rd order despite the fact that these same actuators were previously identified in \cite{hopkins2015embedded} as 2nd order. A lengthy investigation was conducted to identify the source of the additional pole including the quality of linear ballscrew, bearing mount and gear train housing backlash, and even in the electrical force sensing circuit. In the end, the conclusion was that the 3rd pole resulted from age of the components. Despite the fact that the LSEA has an additional pole, the model captures the system response quite well and contains a sufficient bandwidth for control purposes. 

Utilizing the nominal plant model in (\ref{eqn:nominal-plant}), a DOB can be created to provide feedback for dealing with nonlinear behavior such as stiction, backlash, and other unmodelled dynamics. The DOB simply compares the previous input with the expected input of the nominal plant model given the measured output. This input and expected input difference is known as a disturbance, $\hat{d}$, and provides feedback to the following actuator command to make the system behave closer to the nominal plant model. However, the inverse of the nominal plant model in (\ref{eqn:nominal-plant}) is non-causal and therefore a 3rd order butterworth filter, $Q_d$, is added to both the input and output channels to maintain the causality constraint. 
\begin{figure}[!t]
\centering
\includegraphics[width=\linewidth]{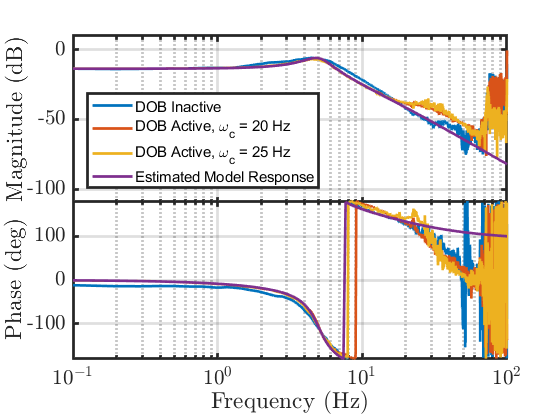}
\caption{Bode plot of plant model from the elastic actuator testbed shows that the system behaves closer to the expected model when the DOB is active. The DOB inactive line is for an $1.75\,\, \rm{A}$ input amplitude.}
\label{fig:dob-bode}
\vspace{-0.5 cm}
\end{figure}
The filter is chosen as
\begin{equation} \label{eqn:dob-filter}
    Q_d(s) = \frac{\omega_c^3}{s^3 + (\sqrt{2} + 1)\omega_cs^2 + (\omega_c^2 + \sqrt{2}\omega_c^2)s + \omega_c^3},
\end{equation}
where $\omega_c$ rad/s is the cut-off frequency. In Figures \ref{fig:dob-bode} and \ref{fig:dob-force-effectiveness}, the responses can be compared of the actuator system with the DOB active and inactive to the nominal plant model. When the DOB is inactive, significant stiction can be seen by the sharp square edges of the force signal in Fig. \ref{fig:dob-force-effectiveness}. This effect creates a phase lag shown by all input amplitudes in Fig. \ref{fig:ol-plant-bode} and when the DOB is inactive in Fig. \ref{fig:dob-bode}. For a filter cutoff frequency of $\omega_c = 20$ and $25\,\, \rm{Hz}$, the DOB provides feedback which removes these nonlinear effects to match the nominal plant model response. The remaining sections utilize a filter cutoff frequency of $\omega_c = 25\,\, \rm{Hz}$. As discussed previously, motor induced vibration affects the system starting around 20 Hz. At this point, the motor-induced vibration begins to cycle through the DOB feedback loop eventually leading to tracking degradation with the model which can be seen in Fig. \ref{fig:dob-bode}.

\begin{figure}[!t]
\centering
\includegraphics[width=\linewidth]{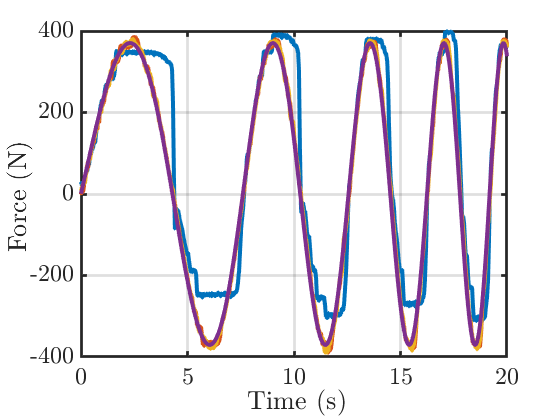}
\caption{DOB effectively removes the nonlinear effects shown here as backlash and provides feedback which makes the system behave very close to the nominal plant model. Note that this plot uses the legend from Fig. \ref{fig:dob-bode}.}
\label{fig:dob-force-effectiveness}
\vspace{-0.5 cm}
\end{figure}

Note that modelling and performance up to this point is still of the elastic actuator testbed and does not include the effects of the structural elasticity from PANDORA's linkages. When transitioning from the actuator testbed to PANDORA, there are significant modelling effects that the DOB is expected to handle. These include the model variation of the structural elasticity of the linkages compared to the LSEA's compliant titanium bar, joint friction, and higher levels of backlash. As discussed in \cite{herron2024pandora, fuge2023design}, 3D printing often leads to greater levels of backlash compared to SM. Unless one is using high-end 3D printers, AM cannot typically compete with the tolerance limits from SM, resulting in non-concentric holes for bearings and joints. Choosing the part orientation while printing also has an effect on joint concentricity where printing a circle in the XY plane results in greater accuracy compared to printing upwards in the XZ or YZ planes. In an effort to lower the backlash, metal components were bolted into the plastic structural components on PANDORA's linkages \cite{herron2023design, fuge2023design}. However, significant backlash is still present on some of the actuator mounting components and within the ball nut/screw interface due to age degradation. 

During testing on PANDORA, the DOB has a tendency to excite backlash leading to a consistent chattering when providing the full disturbance feedback and could lead to instability in certain cases. However, the force controller alone is incapable of dealing with stiction or structural elasticity and is generally unstable without the DOB. As displayed in Figures \ref{fig:joint-space-controller} and \ref{fig:dob-design}, the feedback term, $0 \leq \gamma \leq 1$, is provided to the control engineer to balance the DOB contribution for a specific actuator. Choosing a $\gamma = 0$ means the DOB provides no feedback to the actuator and a $\gamma = 1$ provides full feedback. The $\gamma$ gains for each actuator can be found in Table \ref{tab:joint-controller-gains} and are typically chosen symmetrically with respect to the robot.

\subsection{Actuator Force Controller}

The force controller is a combination of the DOB and PID + FF controller. As discussed in Section \ref{sec:structural-elasticity}, the DOB is responsible for removing nonlinearities such as stiction, backlash, and unmodeled dynamics and is the critical mechanism for handling the structural elasticity. The PID + FF is the feedback-based controller for tracking the desired actuator force command. Without the DOB present, the PID controller is generally unstable due to nonlinearities such as stiction. In addition, the FF term is critical because the system generally needs to provide an input to maintain a stable output. Therefore, the PID + FF and DOB both play a crucial role in tracking the desired force trajectory. 
\begin{figure}[!t]
\centering
\includegraphics[width=\linewidth]{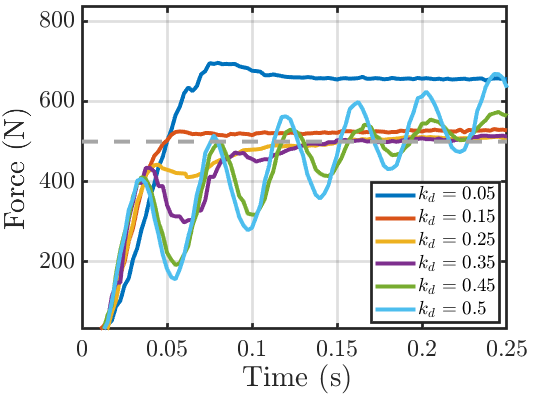}
\vspace{-0.78 cm}
\caption{Force control output on the elastic actuator testbed for a variety of $k_d$ gains with the DOB active at $\gamma = 1$. As the $k_d$ gain increases, the system expectedly becomes more unstable.}
\label{fig:derivative-gain-tuning}
\vspace{-0.5 cm}
\end{figure}
The PID, $C(s)$, can be represented as a causal transfer function in the Laplace domain as
\begin{align}
    C(s)    &= k_p + \frac{1}{s}k_i + \frac{s\lambda}{s + \lambda}k_d, \notag \\
            &= \frac{(k_p + k_d \lambda)s^2 + (k_p\lambda + k_i)s + k_i \lambda}{s^2 + \lambda s},
\end{align}
where $k_p$, $k_i$, and $k_d$ are the proportional, integral, and derivative gains and $\lambda$ is the derivative lower-pass filter coefficient and maintains the causality of the PID transfer function. 
In Fig. \ref{fig:derivative-gain-tuning}, the derivative gain, $k_d$, tuning can be seen for a unit step of $f^d = 500\,\,\rm{N}$ on the elastic actuator testbed. It can be seen that the actuator becomes responsive and unstable as the derivative term increases. Originally, the PID gains were tuned for the elastic actuator testbed and were expected to not require modification due to the model robustness from the DOB. However, during zero impedance and balance testing on PANDORA, it was found that each of the actuators required different PID gains. The PID gains and $\lambda_c$ terms can be found in Table \ref{tab:joint-controller-gains} where $\lambda = 10^{-3}\lambda_c$. 

\subsection{Virtual Impedance}
As displayed in Fig. \ref{fig:joint-space-controller}, the virtual impedance dynamics are designed to act as a linear spring and damper in the actuator space. The desired actuator force, $f^d$ , is passed as the input to the actuator force controller as
\begin{equation} \label{eqn:impedance-force-command}
    f^d = (\bar{q}_{\rm a}^{\rm d} - \hat{q}_{\rm a})\cdot k + (\dot{\bar{q}}_{\rm a}^{\rm d} - \dot{\hat{q}}_{\rm a})\cdot b + f^{\rm d}_{\rm ff},
\end{equation}
where $\hat{q}_{\rm a}$ and $\dot{\hat{q}}_{\rm a}$ are the estimated actuator position and velocity, respectively. The actuator position, $\hat{q}_{\rm a}$, can be estimated from the reduction of the actuator gear transmission and the motor position measurement from the quadrature encoder. The motor velocity is estimated using the QED module's velocity estimator on the LLC's microcontroller which can estimate the actuator velocity using the same gear reduction \cite{herron2023design}. The linear actuator space for various linkages can be better understood using Fig. \ref{fig:lower-body-measurements}. The virtual impedance gains, $k$ and $b$, act as a virtual spring and damper, respectively. These gains have been individually tuned for PANDORA's actuators and provide more stability to the system. The gains for each of the actuators can be found in Table \ref{tab:joint-controller-gains} and are chosen symmetrically with respect to the robot. The FF desired force, $f^d_{ff}$, comes from the kinematic conversion of the FF desired torque, $\tau^{\rm d}_{ff}$, from the WBC. Finally, the $\bar{q}_{\rm a}^{\rm d}$ and $\dot{\bar{q}}_{\rm a}^{\rm d}$ are the desired actuator position and velocity that output from leaky integration of the desired joint accelerations and kinematic conversion. 

\begin{figure}[!t]
\centering
\includegraphics[width=\linewidth]{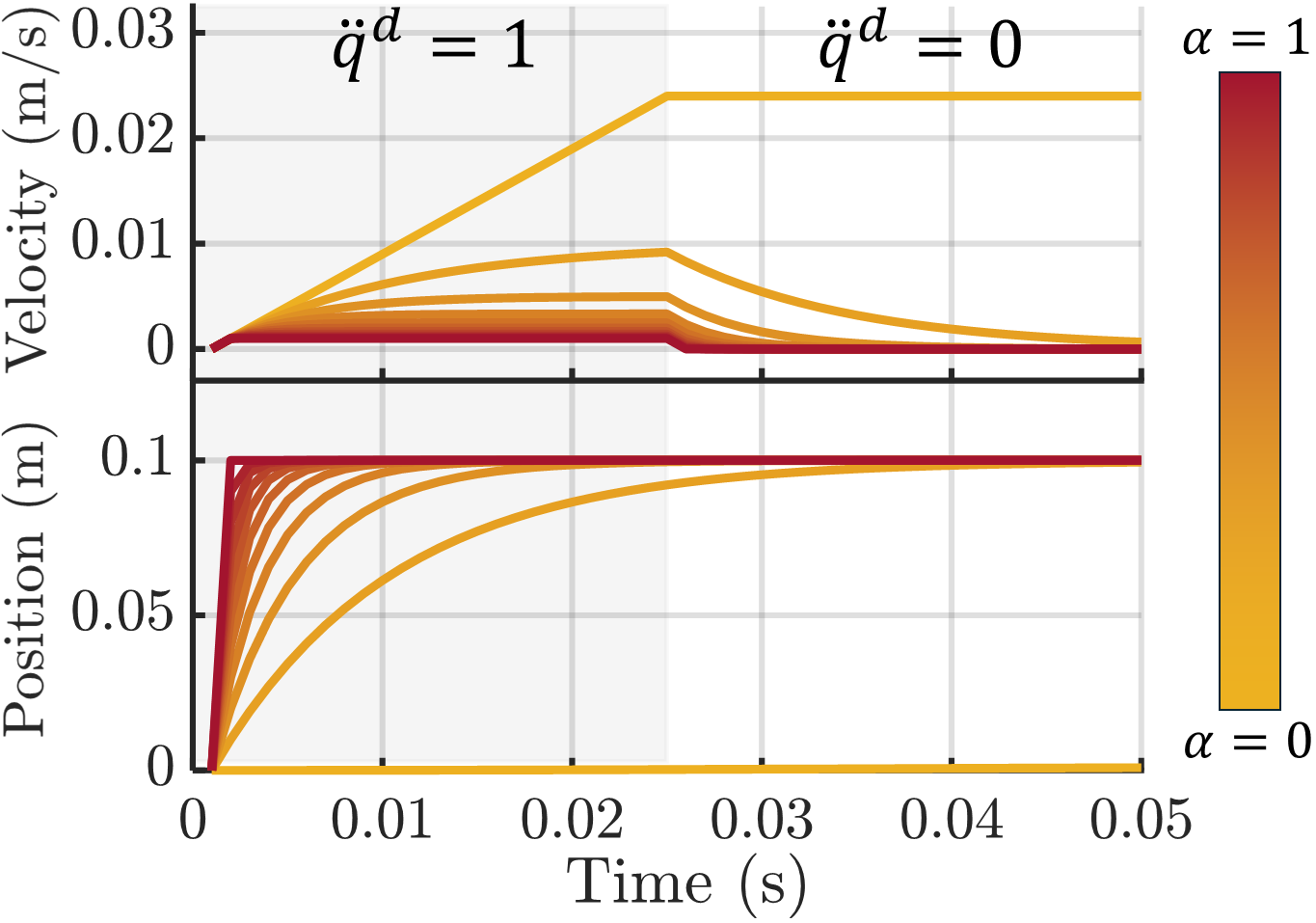}
\vspace{-0.375 cm}
\caption{Leaky integration for a constant input $\ddot{q}^{\rm d} = 1$ from $t = 0$ - $0.025$ and $\ddot{q}^{\rm d} = 0$ from $t = 0.025$ - $0.05$ for various $\alpha = \alpha_v = \alpha_p$ values. The measured position is constant where $q_k = 0.1$ and the step time is $\Delta T = 0.001$ s.}
\label{fig:leaky_integration}
\vspace{-0.5 cm}
\end{figure}

\subsection{Leaky Integration and Kinematics} \label{sec:leaky-integration}
For executing stable behaviors, the virtual impedance controller requires desired position and velocity set points. From the WBC, the desired joint accelerations are obtained where an integration method can be utilized to determine joint position and velocity trajectories. Directly integrating the joint accelerations is subject to integrator windup which eventually will destabilize an actuator. A valuable approach is to \textit{leak} the joint position and velocity setpoints biased toward the current position and a zero velocity. This work utilized the leaky integration method in (C.41) - (C.46) from \cite{jorgensen2020towards} which is simply the combination of an alpha filter and an integrator. For ease of the reader, the leaky integration as a discrete algorithm has been copied below. Note that the subscript $_j$ is previously used in Fig. \ref{fig:joint-space-controller} to denote joint space trajectories for this algorithm, but has been dropped here for mathematical cleanliness. The algorithm can be written as
\begin{align}
    \dot{\bar{q}}_{\rm k+1}^{\rm d} &= \ddot{q}_{\rm k}^{\rm d}\Delta T + (1-\alpha_v) \dot{\bar{q}}_{\rm k}^{\rm d}, \label{eqn:leaked-velocity} \\
    \bar{q}_{\rm k+1}^{\rm d} &= \dot{\bar{q}}_{\rm k}^{\rm d}\Delta T + (1 - \alpha_p)\bar{q}_{\rm k}^{\rm d} + \alpha_p q_{\rm k}, \label{eqn:leaked-position}
\end{align}
where $\ddot{q}_{\rm k}^{\rm d}$ is the $\rm k$th desired joint acceleration from the WBC command, $\dot{\bar{q}}_{\rm k}^{\rm d}$ is the $\rm k^{\text th}$ leaked desired joint velocity, $q_{\rm k}$ is the current joint position measurement, and $\bar{q}_{\rm k}^{\rm d}$ is the $\rm k^{\text th}$ leaked desired joint position, $\Delta T = 0.002$ is the step time, and the subscript $\rm k$ denotes the digital step index. The desired leaked position and velocity are set to track the subsequent, $\rm k+1$, step using the current desired joint acceleration, $\ddot{q}_{\rm k}^{\rm d}$. The $\alpha_v, \alpha_p \in [0, 1]$ are the alpha filter rates of the velocity and position coordinates, respectively, and are chosen as $\alpha_v = \alpha_p = 0.75$ for this work. 

As example of this leaky integration algorithm can be seen in Fig. \ref{fig:leaky_integration} for various $\alpha_v$ and $\alpha_p$ values. For $t = 0$ - $0.025$ s, the acceleration is a constant value where $\ddot{q} = 1\,\, \rm{m/s^2}$ and for $t = 0.025$ - $0.05$ s, the acceleration is $\ddot{q} = 0\,\, \rm{m/s^2}$. 
The step time is set to $\Delta T = 0.001$ s and the measured position is fixed where $q_{\rm k} = 0.1$ m. Notice that the integration windup can be observed when $\alpha = 0$, where after $t = 0.025$ the desired velocity trajectory stays at a constant value. As the $\alpha$ increases, the position converges slower to the fix value of 0.1. Also, it can be seen that the desired velocity in (\ref{eqn:leaked-velocity}) will drop to zero over time if the $\ddot{q}^{\rm d} = 0$ and $\alpha \neq 0$. In the end, when $\alpha_v$ is closer to 1 the system is allowed to change more quickly whereas when $\alpha_v$ is closer to 0 the velocity is more biased to zero. When $\alpha_p$ is closer to 1, the position estimate is more closer to the current position measurement, $q_{\rm k}$, and when $\alpha_p$ is closer to 0, the estimate converges more slowly but is still biased to the current measurement. These details are useful for revealing some side effects and challenges that result from leaky integration which will become important in the following section.
The leaky integration process has been denoted as $L(q_{\rm j}, \ddot{q}_{\rm j}^{\rm d})$ for Fig. \ref{fig:joint-space-controller}, but simply represents (\ref{eqn:leaked-velocity}) and (\ref{eqn:leaked-position}). For further reference regarding this process, an alternative leaky integration approach can be seen in (3.24) from \cite{hopkins2014dynamic}.

After converting the desired joint acceleration into joint position and velocity set points via leaky integration, the joint setpoints are converted into actuator setpoints. As discussed previously and displayed in Fig. \ref{fig:humanoid-control-overview}, each LLC is responsible for controlling an LSEA that drives a 1 DoF joint by itself or a 2 DoF joint with another actuator. The joint space commands are transformed into the actuator space to avoid including the kinematic conversion in the control loop of the lower computation LLC. Since the actuators are either connected in 1 or 2 DoF systems, the joints and actuators are arranged in pairs. For PANDORA's lower body, the pairs are configured as hip roll/yaw, hip and knee pitch, and ankle roll/pitch joints and their driving actuators. The following math is written for a general pair of joints and actuators. The leaky integrated joint positions, $\bar{q}_{\rm j}^{\rm d} \in \mathbb{R}^2$, can be transformed from the joint to the actuator space using the forward kinematics
\begin{equation} \label{eqn:qa_d}
    \bar{q}_{\rm a}^{\rm d} = f_{\rm ja}(\bar{q}_{\rm j}^{\rm d}),
\end{equation}
where $\bar{q}_{\rm a}^{\rm d} \in \mathbb{R}^2$ are the desired actuator positions. Subsequently, the desired actuator velocities can be solved for 
\begin{equation}
    \dot{\bar{q}}_{\rm a}^{\rm d} = \mathbf{J}_{\rm ja}(q_{\rm j})\hspace{-0.025cm} \cdot \hspace{-0.025cm} \dot{\bar{q}}_{\rm j}^{\rm d},
\end{equation}
where $\mathbf{J}_{\rm ja}(q_{\rm j}) \in \mathbb{R}^{2 \times 2}$ is the Jacobian from the joint to actuator space which depends on the current joint measurements $q_{\rm j} \in \mathbb{R}^2$ from the absolute encoders, and $\dot{\bar{q}}_{\rm j}^{\rm d} \in \mathbb{R}^2$ are the desired leaked joint velocities. Finally, the feedforward desired joint torque, $\tau_{\rm ff}^{\rm d} \in \mathbb{R}^2$, is transformed into the actuator space as
\begin{equation} \label{eqn:ff-force}
    f_{\rm ff}^{\rm d} = \mathbf{J}_{\rm ja}^{\rm -T}(q_{\rm j})\hspace{-0.025cm}\cdot \hspace{-0.025cm} \tau_{\rm ff}^{\rm d},
\end{equation}
where $f_{\rm ff}^{\rm d} \in \mathbb{R}^2$ is the feedforward desired actuator force. These desired actuator position, $\bar{q}_{\rm a}^{\rm d}$, velocity, $\dot{\bar{q}}_{\rm a}^{\rm d}$, and force, $f_{\rm ff}^{\rm d}$, trajectories are all composed into a desired actuator force command using (\ref{eqn:impedance-force-command}).

\subsection{Efficient Continuous to Digital Transformation}
Besides various sensors that utilize analog filters to attentuate high frequency noise, most controllers and filters are implemented digitally because of simplicity, flexibility, and effectiveness. Over the course of this work, several controllers and filters were designed as proper transfer functions of up to 8th order. Therefore, a flexible and efficient method of solving for the digital input/output coefficients was necessary. The continuous to discrete representation is completed using the bilinear transform (Tustin's method) technique. 
One can certainly solve for the discrete coefficients offline in a program like MATLAB, but this is inconvenient because it requires the additional step of externally computing the coefficients every time the transfer function is updated. Solving for the discrete coefficients online during initialization makes it easy to test higher order filters and controllers. Therefore, a bilinear transform solver for polynomials is implemented using Horner's method \cite{davies1974bilinear}. This subsection provides a brief overview of this efficient transformation where more details regarding the algorithm can be found in \cite{herron2024software}.

The algorithm presented in \cite{herron2024software} is designed to provide accurate continuous to discrete representation with a low enough computation cost for a microcontroller. This approach can solve for the transformation for any causal transfer function. The general continuous transfer function can be written in the Laplace domain as
\begin{equation}
    H(s) = \frac{Y(s)}{X(s)}= \frac{b_0s^{\rm m} + b_1s^{\rm m-1}+...+b_{\rm m-1}s + b_{\rm m}}{a_0s^{\rm n}+a_1s^{\rm n-1}+...+a_{\rm n-1}s + a_{\rm n}},
\end{equation}
where the output of this algorithm is a set of digital coefficients for the discrete equation
\begin{align} \label{eqn:discrete-eqn} 
    \rm{y}_0 =\, &\hat{b}_1 \rm{y}_1 + \hat{b}_2\rm{y}_2 + ... + \,\hat{b}_{n-1}{y}_{n-1} + \notag\\ 
    \,&\hat{a}_0 \rm{x}_0 + \hat{a}_1\rm{x}_1 + ... +\, \hat{a}_{n-1}\rm{x}_{n-1}, \notag\\
        =\, &\hat{\textbf{b}}^T\textbf{y} + \hat{\textbf{a}}^T \textbf{x},
\end{align}
where $\hat{\textbf{b}}, \textbf{y} \in \mathbb{R}^{\rm n-1 \times 1}$, $\hat{\textbf{b}} = \rm{col}[\hat{b}_1, \hat{b}_2, ... , \hat{b}_{\rm n-1}]$ and $\hat{\textbf{a}}, \textbf{x} \in \mathbb{R}^{\rm n \times 1}$, $\hat{\textbf{a}} = \rm{col}[\hat{a}_0, \hat{a}_1, ... , \hat{a}_{\rm n-1}]$ with $\hat{b}_i, \hat{a}_i \in \mathbb{R}$ are scalars and $n$ is the system order. The vectors $\textbf{x} = \rm{col}[x_0, x_{\rm 1}, ..., x_{\rm n-1}, x_{\rm n}]$ and $\textbf{y} = \rm{col}[y_1, y_{\rm 2}, ..., y_{\rm n-1}]$ represent the input/output ``history" where the subscript represents the loop iteration (eg. $\rm{x}_0$ is the current input, $\rm{x}_1$ is the previous input, $\rm{x}_2$ is the previous previous input, etc.). The bilinear transform method converts continuous time variables, $s$, to discrete time variables, $z$. 
Traditionally, this is solved via substitution of Tustin's Bilinear Transform which is simply a first order approximation of the Taylor series written as 
\begin{equation} \label{eqn:tustins_nominal_transform}
    s = \frac{2}{T}\cdot\frac{z-1}{z+1},
\end{equation}
where $T$ represents the discrete time step in seconds. Using (\ref{eqn:tustins_nominal_transform}) to solve the mapping of $F(s) \rightarrow F(z)$ results in multiple orders of $z$ variables which are difficult to write into an general algorithm for any causal transfer function. In \cite{davies1974bilinear}, an equivalent mapping of $F(s) \rightarrow F(z)$ is suggested for (\ref{eqn:tustins_nominal_transform}) using 
\begin{equation} \label{eqn:tustins_davies_transform}
    \frac{1}{s} = \frac{T}{2}\cdot\big(\frac{2}{z-1} + 1\big).
\end{equation}
Rather than direct substitution of (\ref{eqn:tustins_davies_transform}) into $F(s)$, this method solves for discrete coefficients using the continous transfer function numerator and denominator polynomials
\begin{equation}
    F(s) = \frac{F_n(s)}{F_d(s)},
\end{equation}
where the transformation of $F_n(s) \rightarrow F_n(z)$ and $F_d(s)\rightarrow F_d(z)$ essentially results in the input and output coefficients, respectively.
The sequence of operations executes the mapping of a general variable, $x \rightarrow (x+1)/(x-1)$, as
\begin{align}
    \begin{split}
        F(x)  &\rightarrow F(x+1) \rightarrow F\left(\frac{1}{x} + 1\right) \rightarrow F(\frac{2}{x} + 1) \\ 
        &\rightarrow F\big( \frac{T}{2}\cdot(\frac{2}{z-1} + 1)\big),
    \end{split}
\end{align}
where $F(x)$ simply represents either $F_n(s)$ or $F_d(s)$. As discussed in \cite{davies1974bilinear}, this approach significantly lowers the number of multiplications compares to other approaches and is simple enough to be executed on low-computation hardware. This method is used to solve for the input/output coefficients online during initialization and then used to solve (\ref{eqn:discrete-eqn}) every time step. 

\begin{figure}[!t]
\centering
\includegraphics[width=\linewidth]{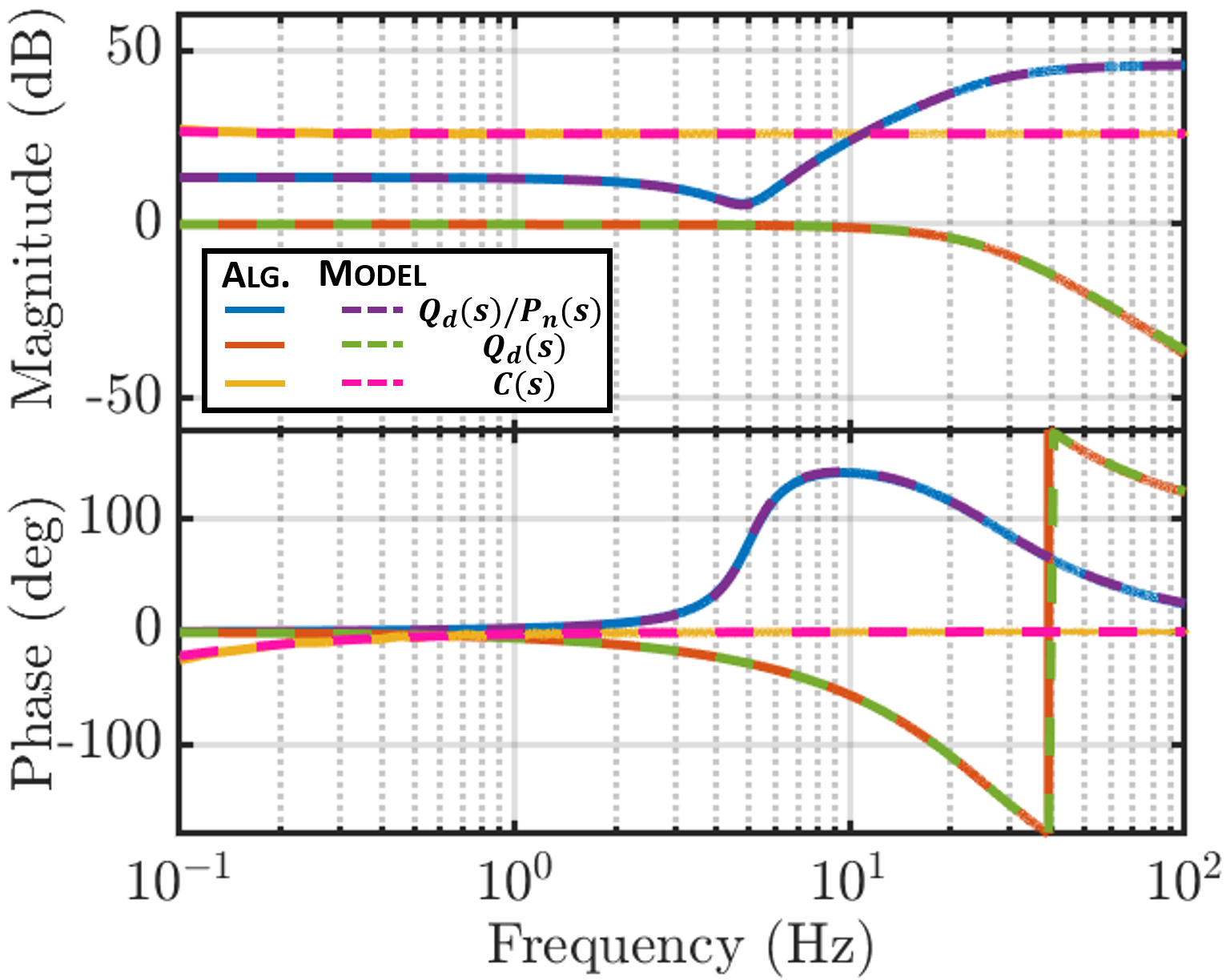}
\caption{Bode plot comparison of the inverse plant model, $Q_d(s)/P_n(s)$, the DOB Filter, $Q_d(s)$, and the PID Controller, $C(s)$ for the algorithm in \cite{herron2024software} compared to the continuous model.}
\label{fig:tustin_exp_model_comparison}
\vspace{-0.5 cm}
\end{figure}

This algorithm is utilized for computing the discrete representation for various continuous transfer function components including the Inverse Plant, $Q_d(s)/P_n(s)$, the DOB Filter, $Q_d(s)$, and the PID Controller, $C(s)$. The result of this transformation is in the form of (\ref{eqn:discrete-eqn}) for each of the transfer function components. Applying a chirp signal through the digital filter, one can generate a bode plot to compare the algorithm's performance to the intended model. In Fig. \ref{fig:tustin_exp_model_comparison}, the algorithm and continuous model responses are compared and demonstrate that the algorithm can correctly capture the model response. An important assumption with digital system implementations is that the timestep must be consistent for the model to behave correctly. This is because the digital coefficients are derived assuming a fixed timestep based on the (\ref{eqn:tustins_nominal_transform}) or (\ref{eqn:tustins_davies_transform}) definitions. This method can be utilized for implementing filters and controllers, and even validating system clock time accuracy. By comparing the experimental data to the intended model's performance, the bode plot of the two should match. Otherwise, the controller time step is not consistent enough and must be updated.

During extensive testing of this algorithm, no limit has been identified on the order or type of transfer function that can be represented besides obeying the causality and Nyquist constraints. However, while first testing the actuator force controller on the LLC, the actuator was unstable even though the bode plots matched the expected behavior. In the end, the system stabilized once of the algorithm was expanded from float to double precision for the components. Furthermore, its been seen that any transfer function above order 10 has very large coefficients where the size and precision of their objects in software should be checked prior to running experiments. More details on the efficient calculation of digital coefficients for continuous systems, worked out examples of this transformation, software psuedocode, and plots of the algorithm response can be found in \cite{herron2024software}. 

\section{Results and Discussion} \label{sec:results_and_discussion}

\begin{figure}[!b]
\vspace{-0.5 cm}
\centering
\includegraphics[width=\linewidth]{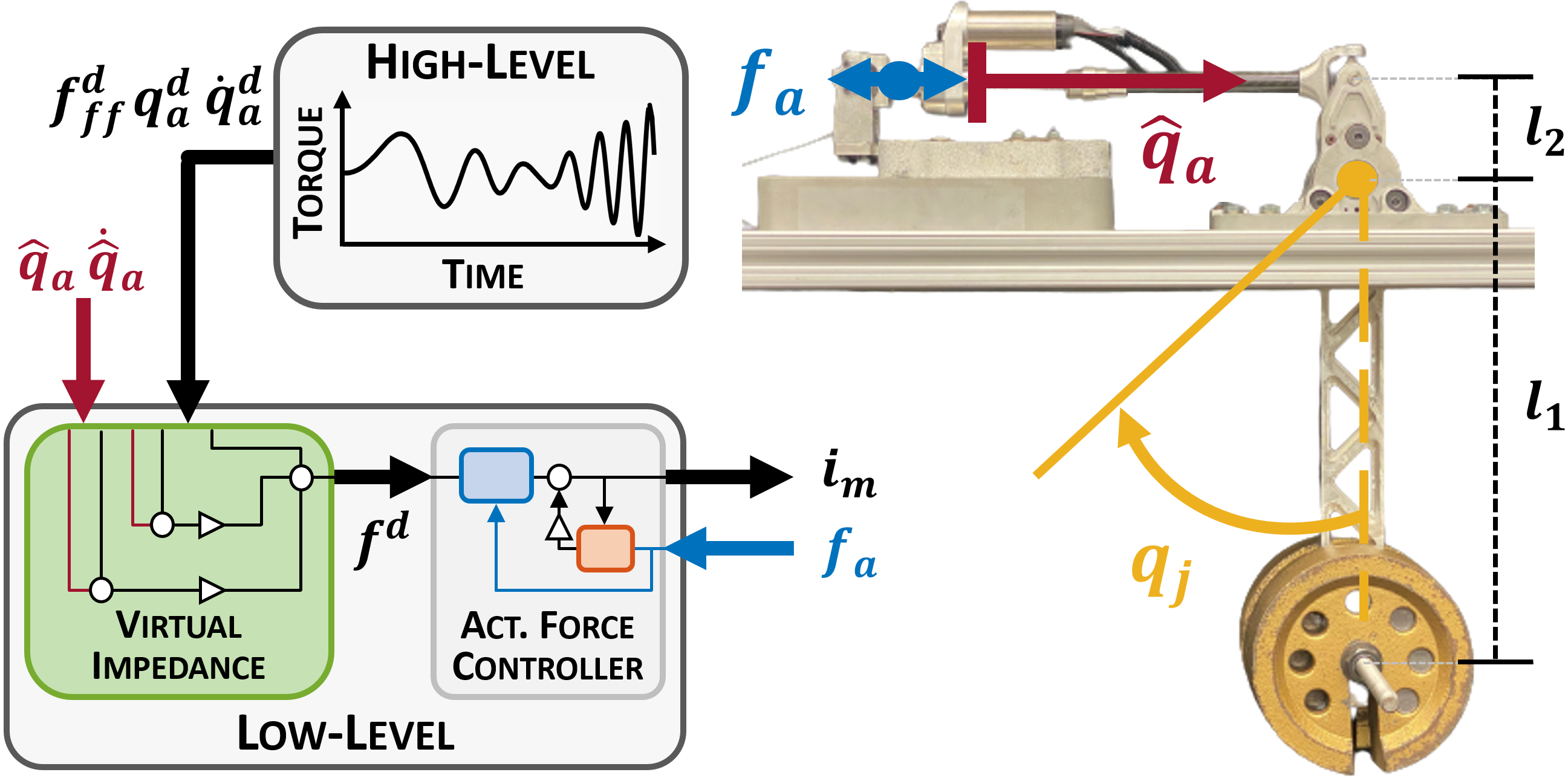}
\caption{Elastic Actuator drives a pendulum with a mass of $m = 10\,\,\rm{kg}$}
\label{fig:pendulum_actuator_testbed}
\vspace{-0.5 cm}
\end{figure}

In this section, the joint-space control method from Section \ref{sec:joint-space-control} is validated using two sets of results. In the 1st experiment, the testbed is unlocked, allowing the elastic actuator to drive a weighted pendulum. These additional pendulum dynamics are not modeled into the actuator force controller and the experiment tests for both scenarios where the DOB is active and inactive. This experiment demonstrates the DOB's robust capabilities in compensating for additional dynamics as a disturbance and supports the claim that this approach can handle the model variation of PANDORA's structurally elastic linkages. The 2nd experiment is tested on PANDORA for performing robust balancing on the 12 DoF lower body.
\begin{figure*}[!t]
\centering
\includegraphics[width=\linewidth]{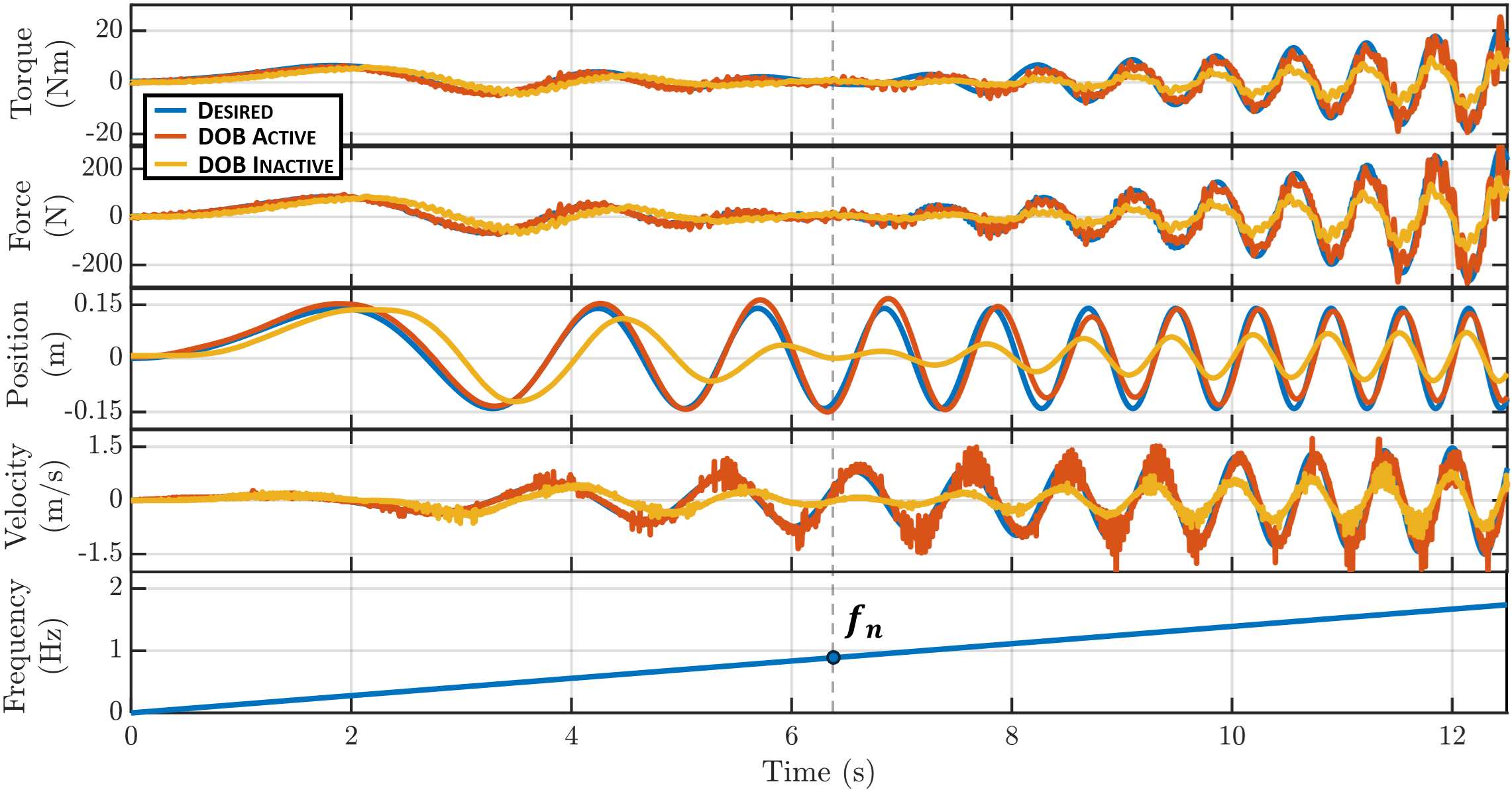}
\caption{In this experiment, the elastic actuator testbed is unlocked and drives a 10 kg pendulum. The controller does not include the model of the pendulum and has a desired trajectory that increases in frequency as a linear chirp signal. At $t = 6.4 s$, the signal crosses the pendulum's natural frequency at $f_n = 0.87$ Hz.}
\label{fig:pendulum_response}
\vspace{-0.5 cm}
\end{figure*}
Unlike the first experiment on the elastic actuator testbed, PANDORA's actuators rely on the link's structural elasticity from the 3D printing compliant material rather than a dedicated elastic component. As discussed in Section \ref{sec:structural-elasticity}, the DOB is modeled on the elastic actuator from the testbed and treats the additional dynamics from the structural elastic components as a disturbance. This experiment validates the robust capabilities of the DOB towards dealing with structural elasticity and the overall joint control approach for humanoid robots.

\subsection{Pendulum Control on Elastic Actuator Testbed}

As displayed in Fig. \ref{fig:pendulum_actuator_testbed}, the actuator mounting point on the right side is unlocked to drive a 1 DoF weighted pendulum with a bob mass of $m = 10\,\,\rm{kg}$. In this experiment, the desired actuator position, velocity, and force are communicated from a high-level controller to the low-level controller at 200 Hz. The desired position is configured as a linear chirp signal which provides tracking performance from 0 - 1.75 Hz. The low-level or joint control algorithm from Section \ref{sec:joint-space-control} is computed within an LLC at 1000 Hz and computes a desired motor current, $i_{\rm m}$, using the estimate actuator position, $\hat{q}_{\rm a}$, velocity, $\dot{\hat{q}}_{\rm a}$, and the measured force, $f_{\rm a}$, in the feedback loop. 

In this case, there is no need to utilize leaky integration to determine desired position and velocity setpoints since they can be analytically derived. In addition, the kinematics from the joint to the actuator space is trivial via the small angles approximation. The desired joint position can be defined as
\begin{equation}
    q_{\rm j}^{\rm d} = A\cdot \sin(\omega_o t^2)
\end{equation}
where $A$ is the amplitude and $\omega_o = 2.75$ is the sinusoidal frequency rate of the linear chirp signal. The subsequent joint velocity and acceleration can be derived using differentiation 
\begin{align}
    \dot{q}^{\rm d}_{\rm j} &= 2A\omega_o t\cos(\omega_o t^2), \\
    \ddot{q}^{\rm d}_{\rm j} &= 2A\omega_o \cos(\omega_o t^2) - 4A \omega_o^2 t^2 \sin(\omega_o t^2).
\end{align}
The joint torque for a weighted pendulum is based on 
\begin{equation} \label{eqn:pendulum-dynamics}
    \tau_{ff}^{\rm d} = m l_1^2\ddot{q}^{\rm d}_{\rm j} + mgl_1 \sin(q_{\rm j}^{\rm d}),
\end{equation}
where $l_1 = 0.33\,\,\rm{m}$ is the length of the pendulum from the pivot point, $g$ is the gravitational constant, and $m$ is the bob mass. The transformation from the joint to the actuator space uses small angle approximation such that $q_{\rm a}^{\rm d} = l_2 q_{\rm j}^{\rm d}$, $\dot{q}_{\rm a}^{\rm d} = l_2 \dot{q}_{\rm j}^{\rm d}$, $\ddot{q}_{\rm a}^{\rm d} = l_2 \ddot{q}_{\rm j}^{\rm d}$, and $f_{\rm ff}^{\rm d} = l_2 \tau_{\rm ff}^{\rm d}$ where $l_2 = 0.07\,\,\rm{m}$ is the pivot offset of the actuator mount displayed in Fig. \ref{fig:pendulum_actuator_testbed}. The desired torque dynamics in (\ref{eqn:pendulum-dynamics}) do not to include a model of the actuator since most full-order models of humanoids typically ignore these dynamics.


As displayed in Fig. \ref{fig:pendulum_response}, the desired actuator trajectories are sent to the LLC and tested for both scenarios where the DOB is active and inactive. The position and velocity plots in Fig. \ref{fig:pendulum_response} are from the actuator space. Immediately, it can be seen that the system response does a better job tracking when the DOB is active, displaying less phase delay and matching the magnitude of the desired position trajectory. At $t = 6.4\,\, \rm{s}$, the system crosses the pendulum's natural frequency at $f_n = \frac{1}{2\pi}\sqrt{g/l_1} = 0.87\,\,\rm{Hz}$. The final plot of Fig. \ref{fig:pendulum_response} is based on the frequency of the position sinusoid 
\begin{equation}
    f = \frac{\omega_o t}{\pi}.
\end{equation}
Around this natural frequency, the actuator is completely unable to track the desired trajectories without the DOB active. For the active DOB, the magnitude is slightly above and below the tracking objective near the natural frequency with very little phase delay. 

\begin{figure*}[!t]
\centering
\includegraphics[width=\linewidth]{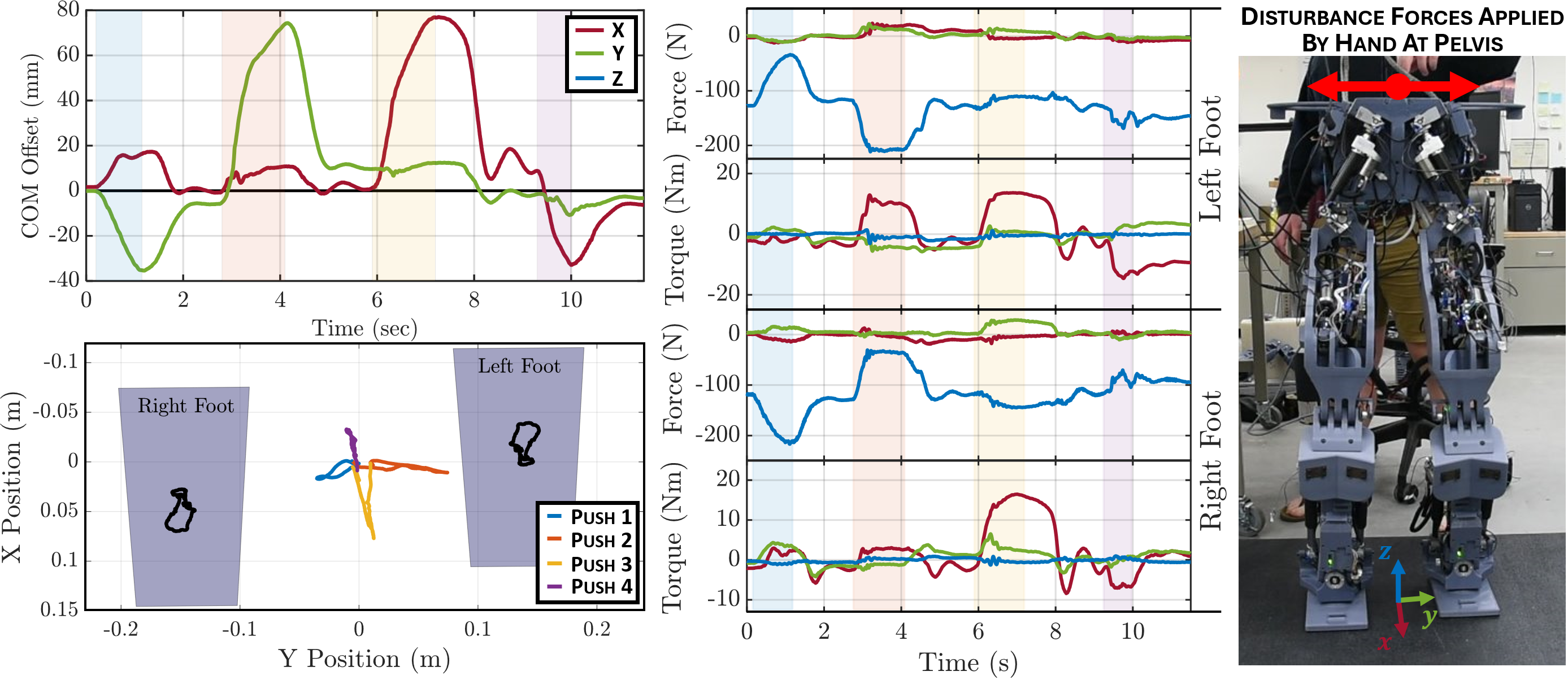}
\caption{Experiment demonstrates PANDORA robustly balancing for 4 pushes applied by an operator in the $\pm \, x$ and $y$ directions. After each push, it can be seen that the CoM returns back to its desired position at the origin. The black lines within the footsteps indicate the estimated motion of the left and right feet from their original position throughout the experiment. This motion likely comes from the structural elasticity and backlash, where the feet positions did not physically change during the test.}
\label{fig:robust-balance-com-cop-grf}
\vspace{-0.5 cm}
\end{figure*}
Notice that despite the position estimate being smooth, the actuator velocity estimate is quite noisy. This noise comes from software differentiation in the LLC's control loop without any additional filtering. Since this experiment, the velocity estimate has been significantly improved by relying on the QED module's differentiation algorithm on the LLC's microcontroller. 




The pendulum experiment demonstrates the DOB's robust ability to handle significant additional dynamics to maintain proper tracking objectives. Without the DOB active, the actuator was unable to track the target trajectories especially near the pendulum's natural frequency. This provides further support that the model variation between the ideal elastic component on the testbed and PANDORA's structurally elastic components can be compensated via the DOB.

\subsection{Robust Balancing on PANDORA}
In this experiment, PANDORA's 12 DoF lower body is controlled via the linear actuators where the elasticity comes from the compliance within the linkages. The hip yaw/roll and ankle pitch/roll joints are 2 DoF where two actuators are simultaneously controlling both joints. The hip and knee pitch joints are 1 DoF and driven by single actuators. The estimates and measurements of the various actuators and joints of the lower body is displayed in Fig. \ref{fig:lower-body-measurements}. In addition, the joint-space controller simply handles 1 and 2 DoF joints through the robot kinematics whereas the impedance and force controller is handled individually on the LLC's for each actuator. PANDORA is controlled using a hierarchical controller similar to Fig. \ref{fig:humanoid-control-overview}. PANDORA's high-level controller handles footstep, CoM, and motion planning, and WBC. The high-level is developed using IHMC's ORS \cite{ihmcORS} and TRS \cite{trecTRS} and runs at 500 Hz on a PC with a 16 core i9 CPU and a GTX 4090 GPU. This PC communicates with the LLCs at 500 Hz, collecting the recent sensor measurements/estimates and providing desired actuator trajectories for control. This experiment tests the full humanoid control approach where the joint-space controller must continuously handle disturbances for the 12 actuators in order to achieve the balancing objective.

\begin{figure*}[!t]
\centering
\includegraphics[width=\linewidth]{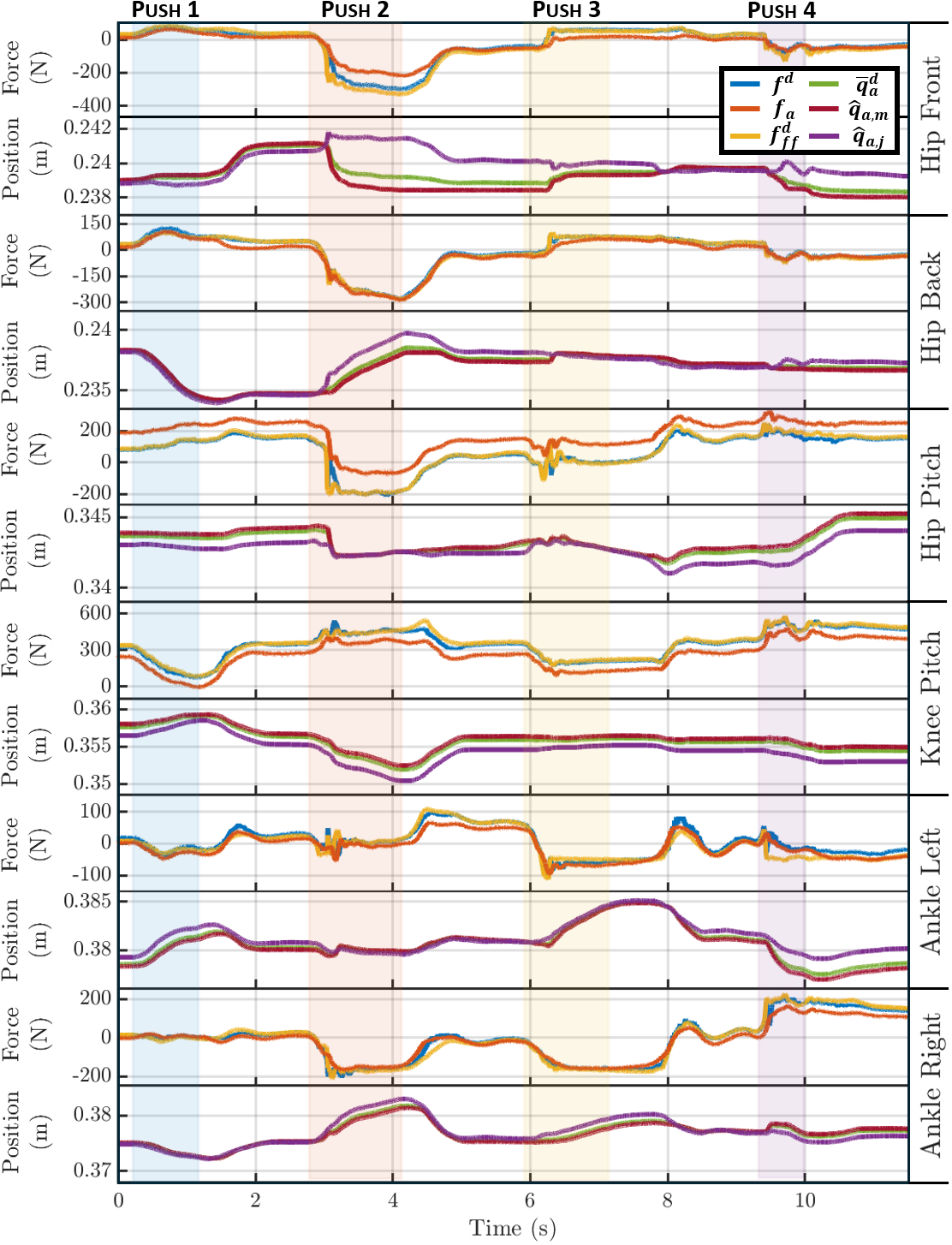}
\caption{Actuator Force and Position tracking for the left leg during the balancing test. The actuator position are estimated using the motor, $\hat{q}_{\rm a,m}$, or joint, $\hat{q}_{\rm a,j}$, encoder where any difference is from backlash or structural elasticity.}
\label{fig:robust-balance-joint-control}
\vspace{-0.5 cm}
\end{figure*}

The CoM planner is controlled via a DCM-based approach which separates CoM dynamics into stable and unstable components. DCM is designed to indirectly stabilize the CoM and results in force vectors that are within the support polygon \cite{mesesan2023unified, griffin2016model, hopkins2015dynamic}. Using the footstep positions, the CoM \textit{follows} the DCM from footstep to footstep to achieve walking behaviors. In this experiment, the footstep positions are held constant and the CoM is driven toward the midpoint of the feet, chosen as the origin.
When a disturbance is applied to the CoM, and it is displaced from the origin, the error feedback creates a linear momentum trajectory that can be tracked by the full-order system to return the COM to the origin. For executing balancing and walking behaviors, the full-order model is controlled using the WBC which requires several tasks including linear momentum, zero feet accelerations, privileged position for the knees, and pelvis orientation and height. By tuning the weights and gains of these tasks, the control engineer can significantly affect the performance and stability of PANDORA. The output of the WBC are the joint accelerations and torques which are the input of the joint-space controller from Fig. \ref{fig:joint-space-controller}.

As displayed in Fig. \ref{fig:robust-balance-com-cop-grf}, an operator applies 4 different disturbance forces to the pelvis in the  $\pm \, x$ and $y$ directions where the high and low-level controllers are responsible for returning the CoM to the midpoint between the feet while maintaining a constant height and pelvis orientation objective. The operator disturbances are estimated to be no more than 50 N using the ground reaction (GR) force measurements at the bottom of the feet. As seen in Fig. \ref{fig:robust-balance-com-cop-grf}, these operator-applied disturbances lead to a CoM offset upwards of 5 cm in the XY direction with approximately 2 seconds between each push.These pushes are ordered by color for Fig. \ref{fig:robust-balance-com-cop-grf} and \ref{fig:robust-balance-joint-control} where blue, orange, yellow, and purple indicate Push 1, 2, 3, and 4, respectively. This experiment can be seen in the supplementive video at \href{www.nothing.com}{videolink.com}.

As displayed in Fig. \ref{fig:robust-balance-com-cop-grf}, the plots on the left side display the CoM offset in time and within the ground plane. For each of the pushes, the CoM is controlled back to the midpoint of the feet in approximately 1 second. The GR wrench measured at the bottom of the feet further support this motion where the first two pushes in the $\pm y$ direction display a shifting in the z-force towards the left and then the right feet. After each push, the z-force returns to approximately 125 N for each foot before being displaced by another disturbance. There's slight variation in the measured torque at the bottom of the feet which is likely due to the fact that the right foot is slightly in front of the left foot.

In addition to CoM offset and GR wrench measurements, Fig. \ref{fig:robust-balance-joint-control} provides the low-level actuator force and position tracking of the left leg where the associated actuator name is given on the right side. The feedforward desired force, $f_{\rm ff}^{\rm d}$ from (\ref{eqn:ff-force}), is directly calculated via kinematics from $\tau_{\rm ff}^{\rm d}$ that outputs from the WBC. The desired force, $f^{\rm d}$ from (\ref{eqn:impedance-force-command}), is the combination of $f_{\rm ff}^{\rm d}$ and the error feedback from the impedance. The desired actuator position, $\bar{q}_{\rm a}^{\rm d}$ from (\ref{eqn:qa_d}), is calculated using leaky integration of the desired joint acceleration from the WBC output and then kinematics to transform from the joint-to-actuator space. The actuator positions can be estimated using either the joint, $\hat{q}_{\rm a,j}$, or the motor, $\hat{q}_{\rm a,m}$, encoders. The transmission dynamics are ignored from the motor-to-actuator space where only the gear reduction is necessary for a smooth estimate. Therefore, the difference between the actuator estimates using the motor and joint encoders is a valuable measure of the elasticity and backlash in the linkages. 

The tracking performance of the left leg is quite similar to the right leg (not shown) where the actuator placement and gains are symmetrically chosen. As displayed in Fig. \ref{fig:robust-balance-joint-control}, the required forces varies dramatically depending on the actuator placement. On PANDORA, the knee and hip pitch actuators consistently measure higher forces when in stance, but the front and back hip actuators can encounter high force variation (beyond 1000 N) for contact transitions during walking. The front/back hip and left/right ankle actuators display the best tracking performance despite being 2 DoF joints. For these joints, both actuators are controlled individually where neither joint objective can be achieved if a single actuator displays poor tracking performance. Notably during the pushes, the hip and ankle actuators display significant deviations in the actuator position estimates. In general, the variation between estimates occur when the actuator forces are higher. For example, the hip front/back actuators during Push 2 display a 3 mm offset in the actuator position estimate continuing into Push 3 even though the force vectors returned to lower values. Since the joint encoder is utilized for determining $f_{\rm ff}^{\rm d}$, this could result in an offset in the desired force which could be difficult for the force sensor to achieve in the given configuration. Had this been the impedance position error, the desired force trajectory would see a 120 N offset in the desired force trajectory. In this case, it is likely that Push 2 created variation in the estimate from structural elasticity of the link along with backlash that did not restore until after the motion from Push 3. This relationship supports the claim that the estimate variation is mostly due to structural elasticity and backlash. Despite this challenge, the actuators and hierarchical controller remain stable and achieve the balancing objective of the robot.

The hip and knee pitch tracking performance is quite important considering the highest force consistently passes through these actuators during stance.
The structural elasticity and backlash can create significant joint measurement error leading to an offset in the calculated actuator force trajectory. From Table \ref{tab:joint-controller-gains}, these actuators are set to have much lower $\gamma$ gains compared to the other actuators. As discussed previously in Section \ref{sec:structural-elasticity}, the $0 \leq \gamma \leq 1$ gain decides how much disturbance to provide as feedback from the DOB where $\gamma = 1$ is full feedback and $\gamma = 0$ acts as if the DOB does not exist. The reason the hip and knee pitch actuators have lower $\gamma$ gains is because their actuator mounting points have significant backlash which the DOB tends to excite. Lowering the $\gamma$ gain too low leads to an inability to remove friction feedback and eventually instability in these actuators where full DOB feedback leads to a consistent "chattering" effect during balancing. In the end, utilizing the $\gamma$ gain is important to maintain actuator stability and is chosen symmetrically on the robot. Despite the challenges and errors in the hip and knee pitch actuators, the supplemental video displays PANDORA's robustness in handling the significant disturbances applied by the operator.

Notice that there is only a small variation between $f^{\rm d}$ and $f_{\rm ff}^{\rm d}$ trajectories where the impedance gains for the actuators are displayed in Table \ref{tab:joint-controller-gains}. During early testing, higher impedance gains more often led to instability in the actuators. This instability did not come from the relative magnitude of tracking error, but from communication frame drop. During testing, communication frames drops occured between the LLC and high-level PC from missed control/communication loop deadlines, poor cable connection, and even garbage collection from the high-level software written in Javascript \cite{ihmcORS}. Unlike C, Java does not require a software engineer to directly handle memory management and automates the removal of unused objects once memory usage exceeds a certain level. During the garbage collection process, all threads are momentarily halted which obviously has significant implications for control. However, this can easily be avoided by instantiating all necessary objects in an initialization process prior to runtime and has been utilized for several leading humanoid robots include Atlas (older generation), NASA's Valkyrie, IHMC's Nadia, and Boardwalk Robotics' Alex \cite{griffin2019footstep, jorgensen2019deploying, chen2023integrable, ackerman2024meet}. Communication delay can have significant implications for all levels of the control loop. In this case, however, impedance can be highly susceptible to instabilities if velocity estimates are made using delayed position values. For example, a velocity estimate can be made using position measurements from incoming frames at a consistent communication rate. If several frames are dropped in a row, the position measurements seem to jump, creating a significant spike in velocity estimate even with filtering. Considering the joint accelerations utilize leaky integration, the desired position and velocity trajectories are close to the current position estimate and zero velocity, respectively. Therefore, the stiffness and damping need to be relatively large in have an impact on control, but this can be dangerous when frame drops occur. To mitigate this issue, significant effort went into achieving consistent communication and control rates \cite{stelmack2024satisfying} and identifying frame drops for momentarily scheduling the impedance gains.

As discussed previously, the structural elasticity can be identified using the difference in the actuator position estimates from the motor and joint encoders. As displayed in Fig. \ref{fig:robust-balance-joint-control}, the difference between the estimates increases as the applied actuator force increases. During a disturbance, PANDORA is still capable of stabilizing its CoM for double stance. As discussed in \cite{herron2024pandora}, this compounding effect can impact CoM and footstep estimation. In Fig. \ref{fig:robust-balance-com-cop-grf}, the black lines within the left and right feet indicate the change in position during the test. As seen in the supplementary video, the footstep positions do not physically change, but the estimate varies by up to 5 cm during the balancing experiment. During transitions from double to single stance, this footstep estimate error increases and leads to instabilities after a few footsteps. Future work will focus on improving this kinematic estimation error via a Kalman filter using the motor encoder, joint encoder, and force sensor measurements. The intent is that the Kalman filter can produce a joint position correction which leads to better kinematic estimation alongside more accurate joint torque to actuator force mapping.

\section{Conclusion} \label{sec:conclusion}
In conclusion, this work provided a joint control method for controlling the linear actuators of the structurally elastic humanoid robot, PANDORA. PANDORA's linkages are intentionally designed to be structural elastic, meaning that the linkages are compliant. Unlike traditional SM methods, additive manufacturing offers lower cost and several options in terms of material. By selecting a compliant material, the linkage can be partially compliant. For PANDORA, the actuator mounting points have been designed to have compliance removing the need for an additional elastic component on the actuator which saves on weight and mechanical complexity. To avoid modeling the elasticity of all linkages, this work utilizes a DOB designed via system identification of an ideal elastic actuator on a testbed.
The DOB provides feedback to remove the model variation from stiction, backlash, and the structural elasticity of PANDORA's linkages. This work provides important details for replicating humanoid robot joint control including leaky integration and efficient digital transformations for implementing the continuous transfer functions from the DOB. In addition, this work features two sets of results where the first is conducted on an elastic actuator testbed and the second is utilizing PANDORA's lower body. In the first experiment, the elastic actuator is unlocked from the testbed and drives a 1 DoF weighted pendulum with a bob mass of 10 kg. The results demonstrate that the actuator is completely incapable of controlling the pendulum without the DOB and validates the DOB's ability to handle model variation. In the second experiment, an operator applies 4 disturbance forces to PANDORA's pelvis where the 12 DoF lower body is controlled to remain balanced. These results validate the overall approach while providing context to current challenges such as torque-to-force transformations, backlash, and state transitions. Future work will focus on improving kinematic accuracy of the feet positions and better handling of stance transitions for achieving continuously stable walking.


%

\ifCLASSOPTIONcaptionsoff
  \newpage
\fi



%
\bibliographystyle{IEEEtran}
\bibliography{bibtex/bib/bibliography} 

%

\begin{IEEEbiography}{Connor Herron}
Biography text here.
\end{IEEEbiography}

\begin{IEEEbiography}{Christian Runyon}
Biography text here.
\end{IEEEbiography}

\begin{IEEEbiography}{Isaac Pressgrove}
Biography text here.
\end{IEEEbiography}

\begin{IEEEbiography}{Benjamin Beiter}
Biography text here.
\end{IEEEbiography}

\begin{IEEEbiography}{Bhaben Kalita}
Biography text here.
\end{IEEEbiography}

\begin{IEEEbiography}{Alexander Leonessa}
Biography text here.
\end{IEEEbiography}




\end{document}